\documentclass[10pt,twocolumn,letterpaper]{article}

\usepackage[pagenumbers]{cvpr} 

\usepackage{graphicx}
\usepackage{amsmath}
\usepackage{amssymb}
\usepackage{booktabs}
\usepackage{cite}

\usepackage{subcaption}
\usepackage[table,xcdraw]{xcolor}
\usepackage{caption}
\usepackage[pagebackref,breaklinks,colorlinks]{hyperref}
\usepackage{pifont}
\newcommand{\cmark}{\ding{51}}%

\usepackage[capitalize]{cleveref}
\crefname{section}{Sec.}{Secs.}
\Crefname{section}{Section}{Sections}
\Crefname{table}{Table}{Tables}
\crefname{table}{Tab.}{Tabs.}

\def\cvprPaperID{5797} 
\def\confName{CVPR}
\def\confYear{2022}

\begin{document}

\title{Evaluating Transformers for Lightweight Action Recognition}

\author{Raivo Koot\\
University of Sheffield\\
{\tt\small raivokoot@gmail.com}
\and
Markus Hennerbichler\\
Speechmatics\\
{\tt\small markushennerbichler@gmail.com}
\and
Haiping Lu\\
University of Sheffield\\
{\tt\small h.lu@sheffield.ac.uk}
}

\maketitle

\begin{abstract}
In video action recognition, transformers consistently reach state-of-the-art accuracy. However, many models are too heavyweight for the average researcher with limited hardware resources.
In this work, we explore the limitations of video transformers for lightweight action recognition.
We benchmark 13 video transformers and baselines across 3 large-scale datasets and 10 hardware devices.
Our study is the first to evaluate the efficiency of action recognition models in depth across multiple devices and train a wide range of video transformers under the same conditions. We categorize current methods into three classes and show that composite transformers that augment convolutional backbones are best at lightweight action recognition, despite lacking accuracy. Meanwhile, attention-only models need more motion modeling capabilities and stand-alone attention block models currently incur too much latency overhead. Our experiments conclude that current video transformers are not yet capable of lightweight action recognition on par with traditional convolutional baselines, and that the previously mentioned shortcomings need to be addressed to bridge this gap. Code to reproduce our experiments will be made publicly available.
\end{abstract}

\section{Introduction}
\label{sec:intro}

Human action recognition is the task of classifying what action is being performed in a video. Due to the high dimensionality of videos, models require massive video datasets to learn robust representations. Additionally, the capacity of models needs to be high enough to effectively model the complex nature of spatiotemporal video data. As a result, prototyping models is subject to long training times, especially on datasets such as Kinetics 400 \cite{k400}, which contains over 90 million RGB images, 60 times more than the popular ImageNet dataset \cite{imagenet}. However, the average researcher has limited hardware resources and requires lightweight models that are feasible to train on as little as a single GPU.

Recent transformer-based models \cite{vivit, timesformer, videoswin, vtn, non-local, mvit, videobert, motionformer} show much promise in dealing with the complexity of video data. They use attention mechanisms  \cite{attention} to suppress redundant information and model long-range interactions in space and time. From the current literature, it is clear which of these models are the most accurate. It is less clear, however, which of these models are best when efficient computation and memory usage is the key constraint. While the usually reported FLOPs metric is hardware agnostic, it does not compare well in practice on current GPUs. In this work, we take a practical approach to answer the question: \textit{How do transformer-based models fare against each other in lightweight action recognition and are they capable of competing with established traditional architectures} \cite{tsm, x3d, movinets}?

\begin{figure}[]
  \centering
  \includegraphics[width=0.80\linewidth]{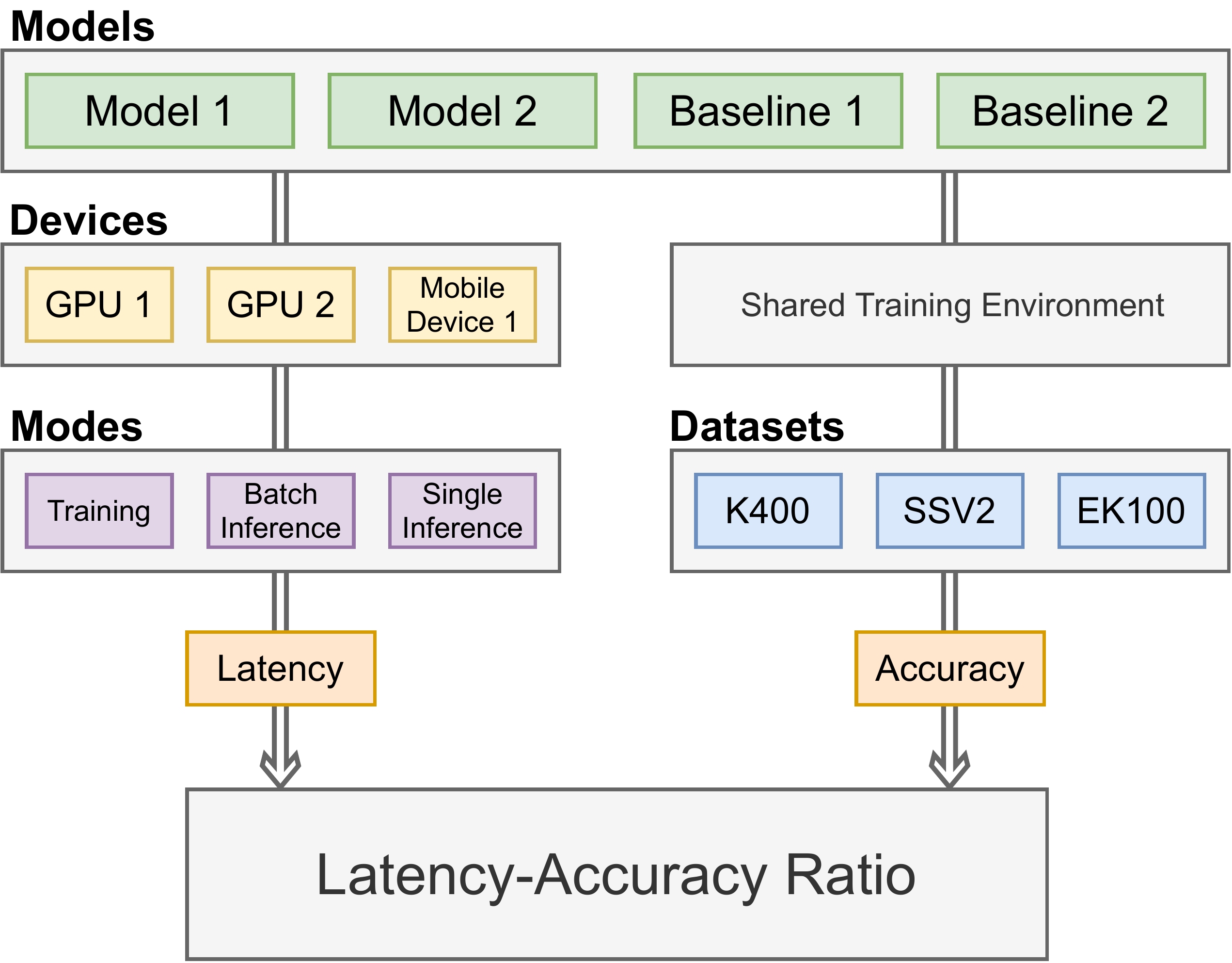}

  \caption[Caption for Figure 1]{Overview of our evaluation methodology (best viewed in color). The latency of 13 models is benchmarked across 10 devices and 3 modes. The accuracy of models is benchmarked under the same training and testing conditions across the Kinetics 400, Something Something V2, and Epic Kitchens 100 datasets. The resulting latency-accuracy ratio measures how capable models are of lightweight action recognition.}
  \label{fig:methodology}
  \vspace{-15pt}
\end{figure}

We gather 13 lightweight instantiations of existing transformer-based models and baselines, and benchmark their efficiency and accuracy. \Cref{fig:methodology} summarizes our methodology. We benchmark the latency of all models on 5 desktop GPUs and 5 mobile devices across 3 computation modes (training, inference, ...). We further evaluate each model under the same training and testing conditions on the large-scale Kinetics 400 \cite{k400}, Something Something V2 \cite{ssv2}, and Epic Kitchens 100 datasets \cite{ek100}. Our study is the first to conduct an in-depth efficiency evaluation of action recognition models across 10 hardware devices. We are also the first to compare a wide range of transformer-based models under the same training and testing conditions on 3 representative datasets.

Our key findings are that 1) attention-only architectures lack motion modeling ability, 2) stand-alone attention block models suffer significant latency overheads, 3) composite transformers that augment convolutional backbones are much more lightweight and accurate, but still fall behind baseline accuracy 4) reducing internal self-attention dimensionality improves latency significantly compared to other approaches, and ultimately 5) current video transformers are not yet as capable of lightweight action recognition as traditional convolutional baselines are.

\section{Related Work}
\label{sec:related_work}
In a recent 2020 study, Zhu et al. \cite{survey} summarize the action recognition performance of 25 recent and historical models. They include an inference latency comparison of seven traditional models, including I3D \cite{i3d}, TSN \cite{tsn}, SlowFast \cite{slowfast}, and R2+1D \cite{r2plus1d}, on a single GPU, when using a batch size of one (single-instance inference). In our work, we focus on 10 modern transformer-based models not included in their study (+3 baselines), and expand efficiency comparisons to a much more in-depth set of benchmarks.

\subsection{Transformers in Action Recognition}
The intersection of transformers \cite{attention} (or simply the self-attention mechanism) and action recognition can be grouped into three distinct categories.
\begin{enumerate}
  \item Stand-alone attention blocks \cite{non-local, tea, w3, tam} that can be inserted into arbitrary existing convolutional models.
    \vspace{-5pt}
  \item Specific compositions of transformer encoders and established existing architectures \cite{vtn, videobert}, such as 2D or 3D CNNs (composite transformers).
    \vspace{-5pt}
  \item Attention-only models, where convolutions are abolished and transformers are the sole building block of the architectures \cite{vivit, timesformer, videoswin, mvit, motionformer, vidtr, stam}.
    \vspace{-5pt}
\end{enumerate}
In each case, the aim is to harness the long-range modeling and feature filtering capability of self-attention.

\subsection{Stand-alone attention blocks}
Given the plethora of current and future convolutional baselines \cite{resnet, mobilenet, effnet, resnext}, the simplest and most future-proof approach is to develop model-agnostic attention blocks that can be used to augment arbitrary traditional models. These studies often show that upgrading a baseline, such as ResNet50 \cite{resnet} or TSM \cite{tsm}, using such a block improves the accuracy while only marginally increasing or even decreasing the number of FLOPs. 

Non-local neural networks propose the non-local block (NL) \cite{non-local}, the basic self-attention mechanism found in transformers. The NL block can be inserted between existing layers of a 2D or 3D CNN and improves long-range spatiotemporal modeling capabilities while increasing the FLOPs requirement. TEA and W3 blocks \cite{tea, w3}, on the other hand, aim to improve efficiency as well and instead use a sigmoid-based excitation attention mechanism, based on similar work for 2D image recognition \cite{squeeze}. The W3 block is inserted between existing layers of 3D CNNs while the TEA block is a spatiotemporal replacement for layers in existing 2D CNNs. These stand-alone blocks have the advantage of being model-agnostic and thus widely applicable, but leave finding the most efficient mixture between baseline and attention block up to experimentation.

\subsection{Composite Transformers}
Other works compose traditional backbones, such as 2D or 3D CNNs, with entire transformer encoders \cite{attention}. We coin these architectures \textit{composite transformers}. The Video Transformer Network (VTN) \cite{vtn} uses a transformer as the final temporal modeling and aggregation component of a Temporal Segment Network (TSN) \cite{tsn}. Similarly, \textit{3D CNN + BERT} \cite{videobert} replaces the final temporal global average pooling module of arbitrary 3D CNNs with a single transformer for improved temporal information aggregation. These model compositions harness the efficiency of their convolutional backbone, and add only little computational overhead (1 to 4 small transformer layers).

\begin{table*}[t]
\scriptsize
\centering
\caption{Overview of the model versions we use in our experiments. When possible, we choose small model configurations, to make training feasible for the average researcher limited to a small number of GPUs.}
\vspace{-8pt}
\label{tab:models}
\begin{tabular}{>{\kern-\tabcolsep}*{3}{lll}<{\kern-\tabcolsep}}
\toprule
Model                       & Version                                                   & Version Name \\ \midrule
\textit{Stand-alone attention block models} \\
NL  \cite{non-local}        & ResNet50 \cite{resnet} backbone augmented with non-local blocks                & NL-R50      \\
TEA     \cite{tea}          & Res2Net50 \cite{res2net} backbone augmented with TEA blocks               & TEA-R50     \\  \midrule
\textit{Composite transformer models} \\
3D CNN + BERT \cite{videobert}  & I3D ResNet50 backbone \cite{i3d} with one temporal transformer layer          & I3D-BERT    \\
VTN     \cite{vtn}          & ResNet50 \cite{resnet} backbone with three temporal transformer layers  & VTN-R50     \\
VTN     \cite{vtn}     & EfficientNet-B0 backbone \cite{effnet} with three temporal transformer layers  & VTN-EFF     \\  \midrule
\textit{Attention-only models} \\
ViViT \cite{vivit} / TimeSformer \cite{timesformer} / VidTr \cite{vidtr}& Base TimeSformer model                                   & TimeSformer  \\
ViViT \cite{vivit} / STAM \cite{stam} / VTN \cite{vtn}                & Factorized encoder (FE) with a DeiT-small backbone \cite{deit} & ViViT-FE      \\
Swin   \cite{videoswin}     & 3D version of Swin-Tiny \cite{swin}                                 & Swin-T       \\
MViT    \cite{mvit}     & Multiscale Vision Transformer Small                       & MViT-S       \\ \midrule
\textit{Non-transformer baselines} \\
TSM     \cite{tsm}      & ResNet50 backbone \cite{resnet} using Temporal Shift Modules            & TSM-R50     \\
X3D      \cite{x3d}    & XS version of X3D based on ResNets \cite{resnet}                             & X3D-XS       \\
MoViNet    \cite{movinets}   & A0 version of MoViNet (minimal version)                & MoViNet-A0   \\
MoViNet    \cite{movinets}   & A2 version of MoViNet (version designed for $224\times224$ inputs)      & MoViNet-A2   \\ \bottomrule
\end{tabular}
\vspace{-10pt}
\end{table*}

\subsection{Attention-Only Models}
Finally, attention-only models explore a radically different approach to standard architectures. They abolish convolutions and their inductive bias altogether, constructing networks consisting only of transformer encoders. While these types of networks seem to require more training data \cite{vit} or stronger regularization \cite{deit}, a seminal work on 2D image recognition by Dosovitskiy et al. \cite{vit} has demonstrated their competitive potential. 

The concurrent works of the Video Vision Transformer (ViViT) \cite{vivit}, TimeSformer \cite{timesformer}, and Video Transformer (VidTr) \cite{vidtr} take the first step in this direction for action recognition, by extending the Vision Transformer (ViT) \cite{vit} for 2D image recognition to videos. To improve computation and memory usage, they each replace standard global self-attention with divided space-time attention. Despite yielding impressive accuracies, these models remain large and slow, as they perform almost no downsampling at all. 

The Motionformer \cite{motionformer} has the same architecture as ViViT, TimeSformer and VidTr but proposes trajectory attention, which biases the self-attention mechanism towards tracking objects through time. While the accuracy of trajectory attention improves upon vanilla and divided space-time attention, it doubles the FLOPs requirement. The authors attempt to alleviate this by using approximations of the attention mechanism instead of full attention, but nevertheless Motionformer also remains large and slow.

Next, ViViT \cite{vivit}, Space Time Attention Model (STAM) \cite{stam}, and a version of VTN \cite{vtn}, explore an approach called factorized encoder, where a 2D ViT is first applied separately to each frame, and a smaller temporal transformer aggregates frame information in the end. This approach improves the computation and memory efficiency of attention-only models, but still only performs little downsampling.

A final approach at improving attention-only efficiency has been explored by the Video Swin Transformer (Swin) \cite{videoswin} and the Multiscale Vision Transformer (MViT) \cite{mvit}. In their works, they aim to create \textit{efficient} attention-only models, by including progressive spatiotemporal downsampling, as is standard in traditional CNNs. Both MViT and Swin use learnt spatiotemporal kernels to downsample the input video with increasing depth of the network. Additionally, MViT uses aggressive downsampling on the internal dimensionality of attention blocks to further reduce the memory and computation requirements. Swin, on the other hand, additionally introduces a cyclic windowed attention mechanism \cite{swin} to reduce the quadratic memory and runtime complexity of self-attention to linear complexity. 

As a result, both Swin and MViT have a much higher computation and memory efficiency than other attention-only models, but at the potential cost of accuracy. An overview of all models discussed is shown in \cref{tab:models}.

\section{Experiments: Efficiency}
While the theoretical FLOPs metric is hardware-agnostic, it is typically used to only measure a model's forward pass complexity and is not a reliable measure when real-world training and inference time is most important. Instead, we measure the latency of all models on 5 desktop GPUs and 5 mobile devices. In this fashion, we answer the following questions:

\begin{itemize}
    \item Which models are fastest during training on desktop GPUs?
    \vspace{-5pt}
    \item Which models are fastest in batch-mode inference on desktop GPUs?
    \vspace{-5pt}
    \item Which models are fastest in single-instance inference on desktop GPUs and mobile devices?
    \vspace{-5pt}
    \item How much video memory do models require during training/inference?
    \vspace{-5pt}
\end{itemize}
Sometimes, models are required to do inference on batches of data, while other times models need to process a single instance at a time using a batch size of one. We coin these terms batch-mode inference and single-instance inference.

\subsection{Models}
While evaluating different models under the same latency budgets is most reliable in comparing the speed-accuracy trade-off of the models, this requires arbitrarily rescaling each individual model architecture to match several common latency budget thresholds. To reduce the scope of our evaluation and because different models achieve their optimal speed-accuracy trade-off at different latency budgets, we instead rely on the expertise of models' authors to provide us with each model's most optimally lightweight configuration. We thus compare the speed-accuracy trade-off of each model, based on these exemplary configurations. Naturally, we exclude large model instantiations, such as TimeSformer-L or ViViT-L \cite{timesformer, vivit}, which are undoubtedly too large for lightweight action recognition.

As efficient baselines, we include the traditional non-transformer models X3D \cite{x3d}, Temporal Shift Module (TSM) \cite{tsm}, and Mobile Video Network (MoViNet) \cite{movinets}. For an apples-to-apples comparison between models, we use the same ResNet50 backbone when possible, as it is most commonly used in public implementations. While this allows for a fairer comparison, this does not necessarily reflect the state-of-the-art of backbone efficiency. To demonstrate the performance achievable when using newer and more efficient backbones, we additionally include a single VTN model that uses an EfficientNet-B0 backbone instead \cite{effnet}. Table \ref{tab:models} shows the model configuration choices we make. The upcoming subsections first compare model efficiency in isolation, after which both efficiency and accuracy are considered jointly in \cref{sec:accuracy}.

\subsection{Experimental Setup}
\label{sec:training_latency_setup}
While different desktop GPUs yield different runtimes and CUDA optimizations for neural network operations can change the latency of networks, we aim to gather a best estimate of the real-world efficiency of models, by measuring the latency of models on a representative set of current GPUs. The GPUs we include are Tesla V100 32GB, Tesla P100 16GB, RTX 3090 24GB, A100 SXM4 40GB, and RTX A6000 48GB. This comprises a mix of data-center and consumer GPUs.

For evaluation on mobile devices, we measure the latency of models on the iPhone 12 Pro Max, iPhone 13 Pro, Google Pixel 5, Samsung Galaxy S20 Ultra, and Samsung Galaxy S9. For the iPhones, models are converted from PyTorch \cite{pytorch} to CoreML \cite{coreml} using the publicly available CoreMLTools. For all Android devices, models are converted from PyTorch to PyTorch Lite \cite{pytorchmobile}. On the Android devices we measure latency on the mobile CPU. On the iPhones we provide separate measurements using their GPU and Apple Neural Engine (ANE) accelerator. Swin and TSM are excluded from mobile benchmarks because their implementations are not mobile compatible.

\vspace{-10pt}

\paragraph{Training}
\label{par:train_setup}
On each GPU, we measure the time it takes each model to perform a single training step using 16 frames, a batch size of 8, and a frame size of $224\times224$. A batch size of 8 fits every model into 24GB of VRAM. Only the Tesla P100 16GB uses a batch size of 4 to avoid memory overflow. We use mixed precision and take the mean latency over 100 runs, excluding 10 warm-up runs. Because each GPU yields different speeds, we report relative latency only, where the fastest model has a latency of $1.0$. Absolute latency, a FLOPs/latency comparison, and additional settings can be found in the supplemental material.

\vspace{-10pt}

\paragraph{Inference}
For batch-mode inference, we use the same setup as for training, but instead measure the forward pass of each model only and set the batch size to 32 to make better use of VRAM. For single-instance inference we do the same and set the batch size to 1. For mobile inference we do the same as for single-instance inference and use full precision.

\begin{figure}[]
  \centering
  \includegraphics[width=0.9\linewidth]{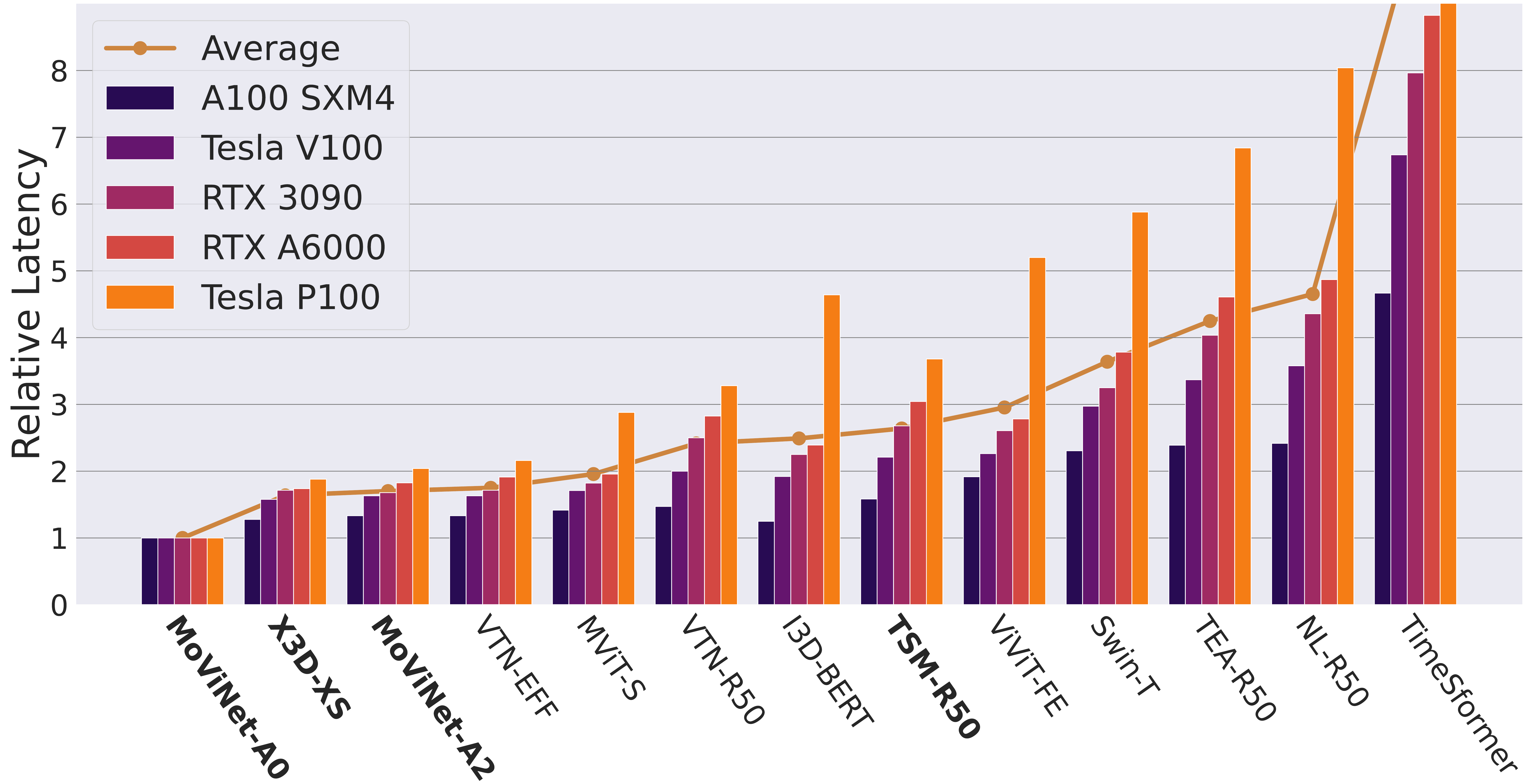}
  \vspace{-8pt}
  \caption[Caption for Figure 1]{Relative training step latency of models on different desktop GPUs (best viewed in color). Baselines are in bold. Additionally, number of epochs until convergence can be found in the supplemental material.}
  \label{fig:train_latency}
\vspace{-15pt}
\end{figure}

\subsection{Training Latency}
\label{sec:training_latency}

\paragraph{Overview} \Cref{fig:train_latency} displays the training latency of each model. While the latency of a model fluctuates across different GPUs, the results remain consistent. It is visible that \textit{no transformer-based model beats the three non-transformer baselines} MoViNet-A0, X3D-XS, and MoViNet-A2, which sit at the very left. The baselines range from a latency of $1.0$ to $2.0$, while most transformer-based models fall between $2.0$ and $5.0$. Even Swin-T, which improves the efficiency of standard attention-only architectures by using progressive downsampling and linear attention complexity, has an average latency that is $3.64\times$ higher than the fastest baseline.

\vspace{-10pt}

\paragraph{GPU Comparison} From left to right, it is visible that the latency on some GPUs explodes more than on others. The maximum latency on the A100, Tesla V100, RTX 3090, RTX A6000, and Tesla P100 is $4.67$, $6.74$, $7.96$, $8.83$, and $22.64$. This means for some GPUs, those where latency jumps up quickly, choosing lightweight models is even more important, because the overall training time can quickly become infeasible with high latency.

\vspace{-10pt}

\paragraph{Stand-Alone Attention Blocks} Similarly, TEA-R50 and NL-R50 show that stand-alone attention block models are some of the slowest at $4.25$ and $4.65$ average latency, respectively. They are about twice as slow as the other models, TSM-R50, VTN-R50, and I3D-BERT, that use the same ResNet50 backbone. So while being model-agnostic, naive \textit{application of these stand-alone attention blocks yields a considerable latency overhead}.

\vspace{-10pt}

\paragraph{Attention-Only Models} In the meantime, the attention-only MViT-S greatly outperforms the other attention-only models ViViT-FE, Swin-T, and TimeSformer, as it achieves an average latency of $1.96$, compared to $2.95$, $3.64$ and $10.2$, respectively. This shows that \textit{attention-only models, including those using factorized encoders (ViViT-FE) and progressive downampling (Swin-T), are far slower than the convolutional baselines}, unless the internal attention dimensionality is aggressively downsampled as in MViT-S.

\vspace{-10pt}

\paragraph{Backbones \& Baselines} Furthermore, VTN-EFF, which uses an EfficientNet-B0 backbone instead of ResNet50, reduces the average latency of VTN-R50 by $27\%$ from $2.42$ to $1.75$, emphasizing the benefit of using more recent backbones. VTN-EFF's latency of $1.75$ is the only transformer-based model that comes close to the baseline latency $1.70$ of MoViNet-A2. Whether MoViNet-A2 and the other baselines show similar superiority in accuracy, however, will be explored later in \cref{sec:accuracy}. Noteworthy is that the baseline TSM-R50 with an average latency of $2.64$ is easily outperformed by the transformer-based models I3D-BERT (same backbone), VTN-R50 (same backbone), and MViT-S with respective latency of $2.49$, $2.42$, and $1.96$.

\vspace{-10pt}

\paragraph{Spread of Latency} Overall, there is a large difference between the maximum average latency ($10.2$) and the minimum ($1.0$). While some transformer-based models like VTN-EFF ($1.75$), MViT-S ($1.96$), VTN-R50 ($2.42$), and I3D-BERT ($2.49$) are able to be much faster than the others ($2.95$, $3.64$, $4.25$, $4.65$, $10.2$), they are still \textit{considerably slower to train than the most lightweight non-transformer baseline} MoViNet-A0 ($1.00$).

\begin{figure}[]
  \centering
  \includegraphics[width=0.9\linewidth]{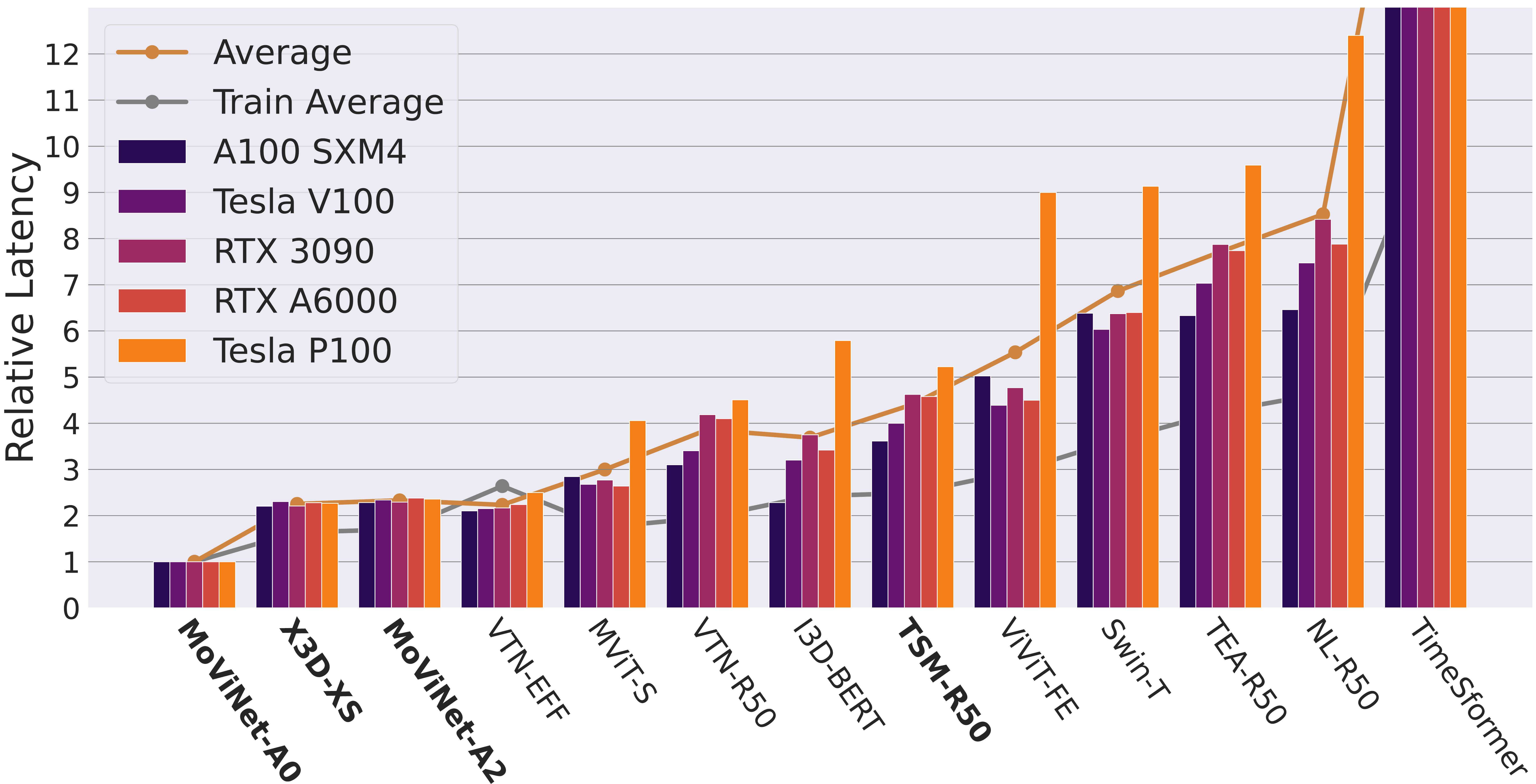}
  \vspace{-8pt}
  \caption[Caption for Figure 1]{Relative \textit{batch-mode} inference latency of models on different desktop GPUs (Best viewed in color). Baselines are in bold.}
  \label{fig:bm_inference_latency}
\vspace{-13pt}
\end{figure}

\subsection{Batch-Mode Inference Latency}
\label{sec:bm_inference}
Batch-mode inference latency results in \cref{fig:bm_inference_latency} are mostly consistent with training latency. However, there are several key differences. While the maximum average training latency is $10.2$, the maximum batch-mode inference latency is $20.1$. While most transformer-based models fall into a latency range of $2.0$ and $5.0$ during training, they fall between $2.0$ and $9.0$ during batch-mode inference. We see a \textit{much larger increase in latency between models during batch-mode inference than during training}.

\begin{figure}[]
  \centering
  \includegraphics[width=0.85\linewidth]{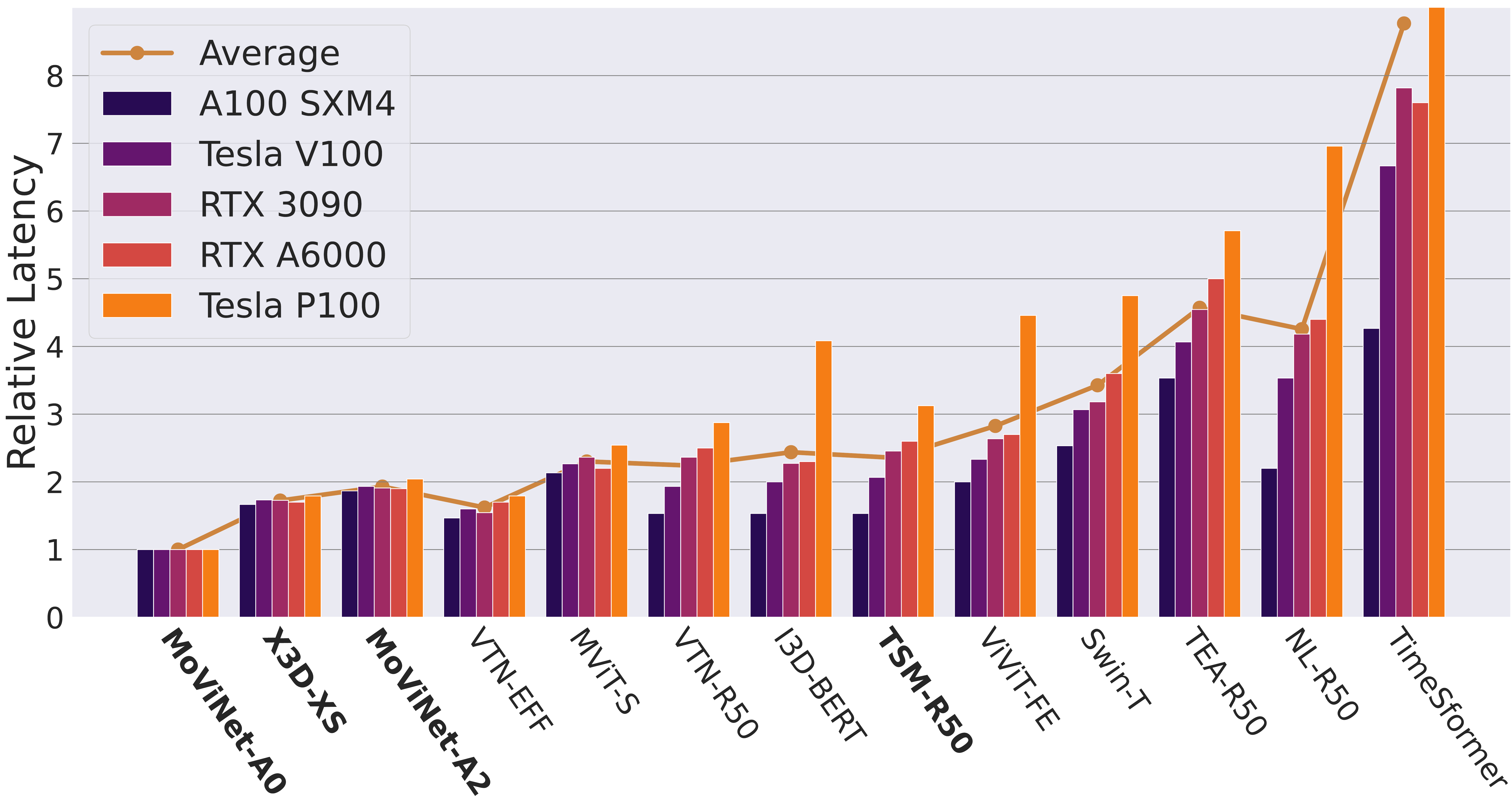}
  \vspace{-8pt}
  \caption[Caption for Figure 1]{Relative \textit{single-instance} inference latency of models on different desktop GPUs (Best viewed in color). Baselines are in bold.}
  \label{fig:si_inference_latency}
\vspace{-10pt}
\end{figure}

\subsection{Single-Instance Inference Latency} 
Single-instance inference latency is shown in \cref{fig:si_inference_latency} and is mostly consistent with training latency and batch-mode inference latency. Compared to batch-mode inference, \textit{single-instance inference latency varies far less} (maximum latency $8.77$ compared to $20.10$). In fact, it even varies slightly less than during training. During single-instance inference most transformer-based models fall into latencies between $1.5$ and $4.5$, compared to $2.0$ and $5.0$ during training. VTN-EFF even surpasses the latency of the X3D-XS and MoViNet-A2 baselines, reaching an average latency of $1.62$ compared to $1.72$ and $1.92$, respectively. Models based on ResNet50 with similar latency to VTN-R50 ($2.24$), namely TSM-R50 ($2.36$) and I3D-BERT ($2.44$), could thus potentially reach a latency similarly favorable to that of VTN-EFF ($1.62$) if a modern EfficientNet backbone was used instead.


\begin{figure}
    \centering
    \begin{subfigure}[b]{0.32\linewidth}
        \centering
        \includegraphics[width=1.0\linewidth]{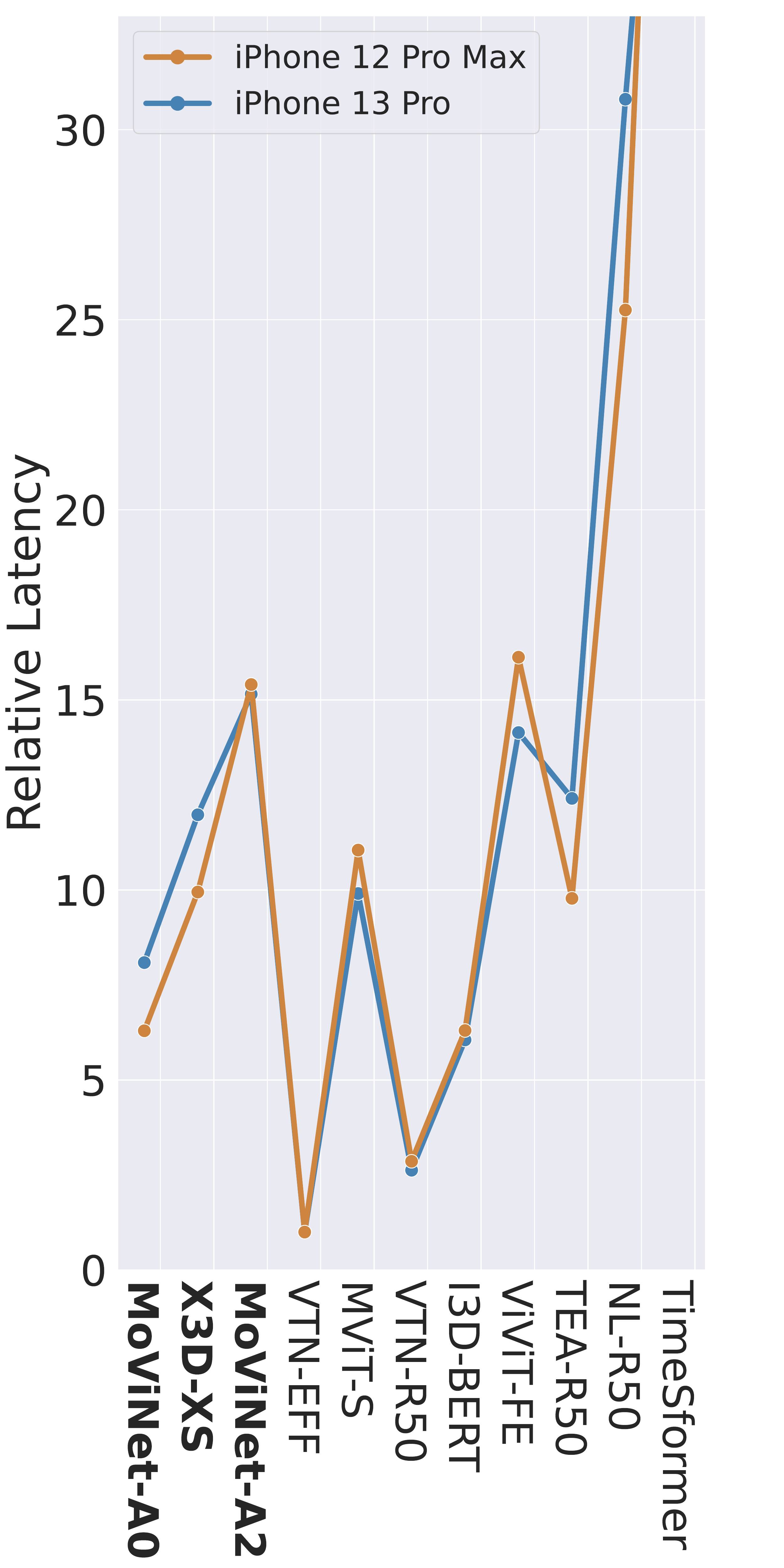}
        \label{fig:ane_mobile}
        \vspace{-11pt}
        \caption{ANE}
    \end{subfigure}
    \hfill
    \begin{subfigure}[b]{0.32\linewidth}
        \centering
        \includegraphics[width=1.0\linewidth]{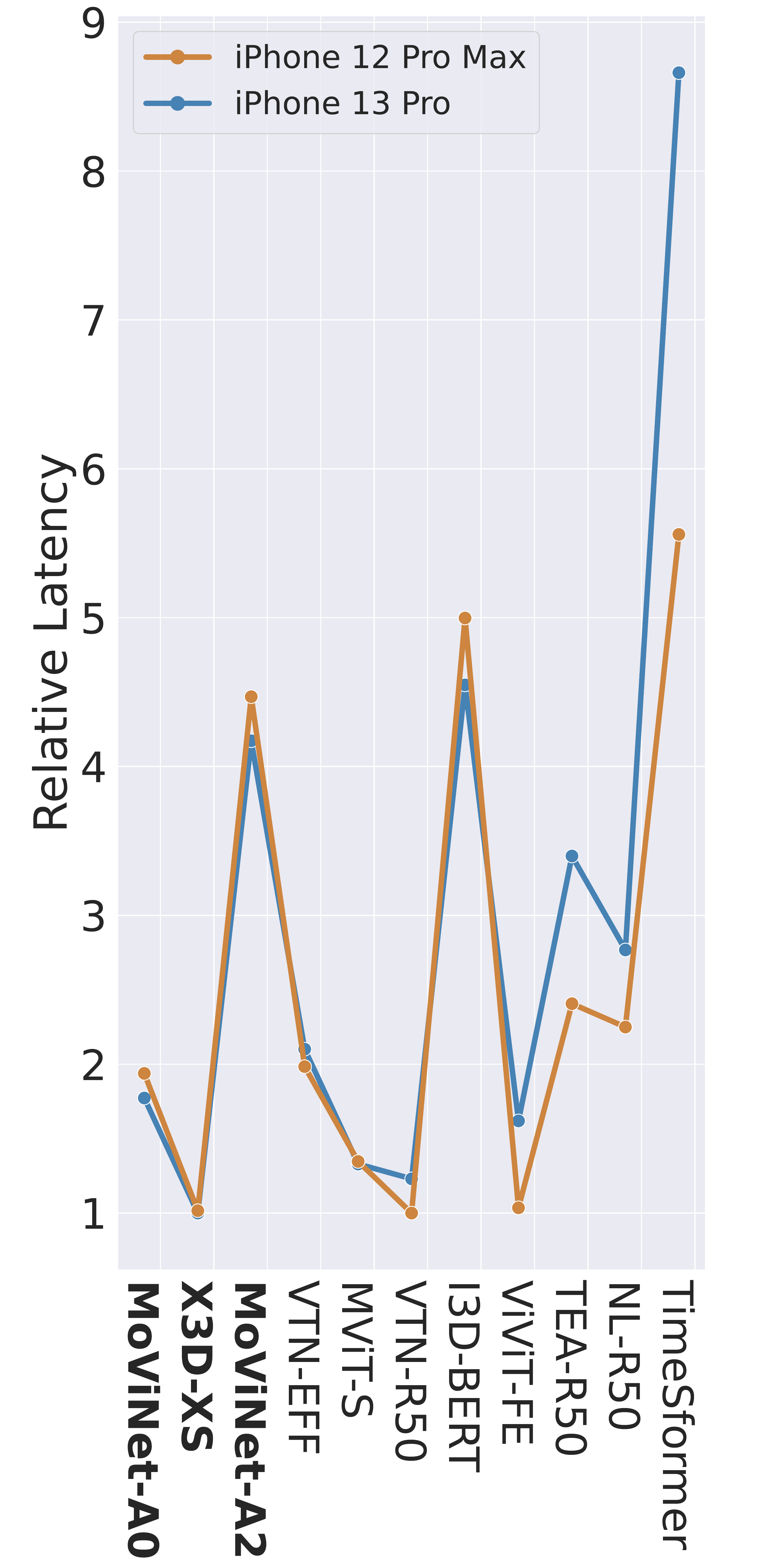}
        \label{fig:gpu_mobile}
        \vspace{-11pt}
        \caption{GPU}
    \end{subfigure}
    \hfill
    \begin{subfigure}[b]{0.32\linewidth}
        \centering
        \includegraphics[width=1.0\linewidth]{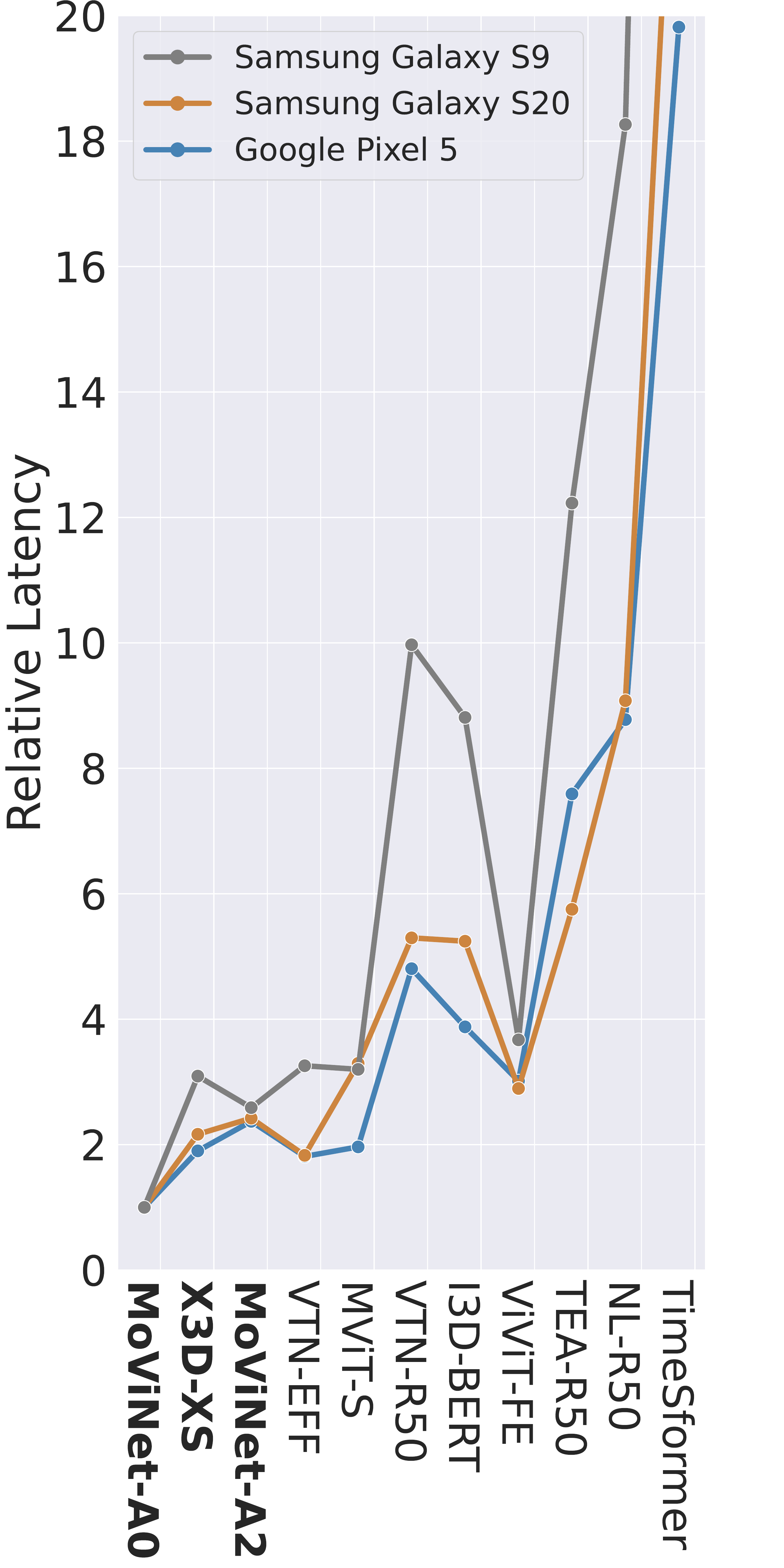}
        \label{fig:cpu_mobile}
        \vspace{-11pt}
        \caption{CPU}
    \end{subfigure}
    \vspace{-8pt}
    \caption{Relative \textit{single-instance mobile} inference latency on the CPU, GPU, and ANE accelerator. Note the different y-axis scales. Latency is significantly less consistent on mobile devices than on desktop devices.}
    \label{fig:mobile_latency}
    \vspace{-13pt}
\end{figure}

\subsection{Inference on Mobile Devices}
Mobile inference results in \cref{fig:mobile_latency} show extreme fluctuations compared to desktop inference. On ANE, GPU, and CPU there is a 57\texttimes, 9\texttimes, and 50\texttimes\, difference between the fastest and the slowest model, respectively. While latency across the three previous desktop benchmarks in \cref{fig:train_latency,,fig:bm_inference_latency,,fig:si_inference_latency} consistently shows the same increasing trend line, mobile latency is not consistent with this. Some models see much better acceleration on mobile than others. This is because some neural network operations have far better mobile optimizations than others. On ANE, both composite transformer VTN models see much better acceleration than any other model. On Android CPUs, ResNet50-based models (VTN-R50, I3D-BERT, TEA-R50, and NL-R50) have much higher latency than the other models. Older iPhones, which do not have an ANE chip use the GPU, where a mix of baselines and transformer-based models in the range of 1.0 and 2.0 (MoViNet-A0, X3D-XS, MViT-S, VTN-R50, ViVIT-FE) perform similarly. Overall, much greater care in choosing efficient models must be taken on mobile devices, and we leave further and more comprehensive analysis of this subject for future work.

\begin{table}[]
\centering
\caption{Comparison of the video memory (VRAM) requirement of each model during training, batch-mode (BM) inference, and single-instance (SI) inference. The same respective input shapes are used as in the corresponding latency benchmarks. Latency for each model is reported as the average across devices.}
\vspace{-6pt}
\label{tab:vram}
\resizebox{0.90\linewidth}{!}{%
\begin{tabular}{@{}lrrrrrc@{}}
\toprule
            & \multicolumn{2}{c}{Training}                         & \multicolumn{2}{c}{BM Inference}                          & \multicolumn{2}{c}{SI Inference}                                             \\ \cmidrule(lr){2-3} \cmidrule(lr){4-5} \cmidrule(lr){6-7}  
Model       & \multicolumn{1}{c}{Latency} & \multicolumn{1}{c}{VRAM} & \multicolumn{1}{c}{Latency} & \multicolumn{1}{c}{VRAM} & \multicolumn{1}{c}{Latency} & \multicolumn{1}{c}{VRAM} \\ \midrule
TimeSformer \cite{timesformer} &   10.20       &   22.2G       &   20.10       &   8.2G      &   8.77        &   2.8G                \\
NL-R50   \cite{non-local}   &   4.65       &   19.5G       &   8.53        &   17.2G     &   4.25        &   2.7G            \\
TEA-R50   \cite{tea} &   4.25       &   12.7G       &   7.71        &   8.8G      &   4.57        &   2.4G                  \\
Swin-T   \cite{videoswin}  &   3.64       &   14.0G       &   6.87        &   8.9G      &   3.43        &   2.5G                   \\
ViViT-FE  \cite{vivit} &   2.95       &   9.9G        &   5.54        &   4.6G      &   2.83        &   2.3G                    \\
I3D-BERT \cite{videobert} &   2.49       &   6.8G        &   3.69        &   4.6G      &   2.44        &   2.4G               \\
VTN-R50  \cite{vtn} &   2.42       &   8.9G        &   3.86        &   7.6G      &   2.24        &   2.4G                   \\
MViT-S  \cite{mvit}  &   1.96       &\textbf{6.0G}  &   3.00        &\textbf{3.5G}&   2.30        &   2.3G                   \\
VTN-EFF \cite{vtn}  &\textbf{1.75} &   9.3G        &\textbf{2.23}  &   6.1G      & \textbf{1.62} &\textbf{2.3G}             \\ 
\midrule
\multicolumn{7}{@{}l}{\textit{non-transformer baselines}}               \\
TSM-R50  \cite{tsm} &   2.64    &   10.3G       &   4.41        &   7.6G           &   2.36        &   2.4G                \\
MoViNet-A2 \cite{movinets} &   1.70    &   8.3G        &   2.33        &   4.8G           &   1.93        &   2.2G                \\
X3D-XS  \cite{x3d}  &   1.64    &   6.7G        &   2.25        &   5.0G           &   1.72        &   2.2G                \\
MoViNet-A0 \cite{movinets} &\textbf{1.00}&\textbf{4.9G}&\textbf{1.00}  & \textbf{3.7G}    & \textbf{1.00} &\textbf{2.2G}         \\
\bottomrule
\end{tabular}
}
\vspace{-9pt}
\end{table}

\vspace{-3pt}
\subsection{VRAM Requirements}
\vspace{-3pt}
Besides latency, the VRAM requirement of a model is one of the biggest constraints during training, dictating which set of models can and can not be trained on a given GPU. While high-tier datacenter GPUs often have 32 or 48GB of VRAM, many common GPUs only have 24GB, 16GB, 12GB, or even less VRAM, seriously restricting the set of usable models. \Cref{tab:vram} gives an overview of how much VRAM each model requires. The table also includes the average latencies from \cref{fig:train_latency,,fig:bm_inference_latency,,fig:si_inference_latency}.

From the training VRAM, which is measured on a batch size of 8, it becomes obvious that many of the models, such as the baselines and the efficient transformer-based models, have a small enough VRAM footprint, such that they can even be trained using a batch size of 16 or 32 on an RTX 3090 with 24GB of VRAM, for example. On the other hand, TimeSformer, NL-R50, Swin-T, and TEA-R50 require too much VRAM (22.GB, 19.5GB, 14.0GB, 12.7GB) to use larger batch sizes during training. In general, the results show that, during training, models with high latency also consistently require much larger amounts of VRAM. On the contrary, the VRAM requirement during single-instance inference does not vary much at all.

\begin{table}[]
\centering
\scriptsize
\caption{Overview of our three benchmark datasets \cite{k400, ssv2, ek100}. The EK100 dataset additionally has 97 and 289 separate verb and noun labels, respectively, which combine to form the action label.}
\label{tab:datasets}
\vspace{-6pt}
\begin{tabular}{@{}lrrrr@{}}
\toprule
Dataset & \multicolumn{1}{c}{\begin{tabular}[c]{@{}c@{}}Training\\ Clips\end{tabular}} & \multicolumn{1}{c}{\begin{tabular}[c]{@{}c@{}}Validation\\ Clips\end{tabular}} & \multicolumn{1}{c}{\begin{tabular}[c]{@{}c@{}}Action\\ Classes\end{tabular}} & \multicolumn{1}{c}{\begin{tabular}[c]{@{}c@{}}Avg. Clip\\ Length (s)\end{tabular}} \\ \midrule
EK100   & 67,217                                                                       & 9,668                                                                          & 3,568                                                                        & $3.1\pm5.4$                                                                        \\
SSV2    & 168,913                                                                      & 24,777                                                                         & 174                                                                          & $3.8\pm1.0$                                                                        \\
K400    & 246,535                                                                      & 19,907                                                                         & 400                                                                          & $10.0\pm0.0$                                                                       \\ \bottomrule
\end{tabular}
\vspace{-10pt}
\end{table}

\begin{table*}[t]
\centering
\scriptsize
\caption{Comparison of models across three datasets. Backbone and Parameter columns specified in \cref{tab:k400}  are the same across all three datasets. IN-1K, and K400 denote ImageNet-1K and Kinetics 400 pretraining, respectively. Latency denotes average training latency (see \cref{tab:vram}). The included model versions are as specified in \cref{tab:models}. Rows are sorted by Top 1 action accuracy. K400 models trained using a more sophisticated training and evaluation strategy from scratch are {\color[HTML]{9B9B9B}de-emphasized}.}
\vspace{-8pt}
\label{tab:accuracy}

\begin{subtable}[t]{.40\linewidth}
    \centering
    \caption{Kinetics 400}
    \setlength{\tabcolsep}{3pt} %
    \resizebox{!}{58pt}{%
    \begin{tabular}{@{}llcccrr@{}}
        \toprule
        Model          & Backbone      & Pretrain   & Top 1         & Top 5         & Latency       & Parameters \\ \midrule
        I3D-BERT \cite{videobert}  & ResNet50 \cite{resnet} & IN-1K & 64.8    &  85.6  & 2.49          &50.6M        \\
        TEA-R50 \cite{tea}   & Res2Net50 \cite{res2net} & IN-1K     & 66.0    &  87.0  & 4.25          &  24.9M  \\
        NL-R50  \cite{non-local}   & ResNet50 \cite{resnet} & IN-1K & 66.9    &  87.7  & 4.65          & 31.7M \\
        VTN-R50 \cite{vtn}  & ResNet50  \cite{resnet} & IN-1K       & 67.2    &  87.4  & 2.42          & 26.5M   \\
        VTN-EFF \cite{vtn}  & Effnet-B0 \cite{effnet} & IN-1K       & 68.2    &  88.1  & \textbf{1.75} &  6.8M   \\
        ViViT-FE \cite{vivit} & DeiT-small \cite{deit} & IN-1K      & 68.8    &  87.7  & 2.95  &  25.4M    \\
        Swin-T  \cite{videoswin}  & Swin-Tiny  \cite{swin} & IN-1K  & 69.8    &  88.7  & 3.64  &  28.0M     \\
        TimeSformer \cite{timesformer}  & DeiT-base \cite{deit} & IN-1K      & \textbf{74.5} &  \textbf{91.5} & 10.20    & 121.0M   \\  \midrule
        \multicolumn{4}{@{}l}{\textit{non-transformer baselines}}              \\
        \color[HTML]{9B9B9B} MoViNet-A0  \cite{movinets} &               &            & \color[HTML]{9B9B9B}65.8          & \color[HTML]{9B9B9B}87.4          & \color[HTML]{9B9B9B}1.00 & \color[HTML]{9B9B9B} 2.7M  \\
        \color[HTML]{9B9B9B}X3D-XS \cite{x3d}& \color[HTML]{9B9B9B}ResNets \cite{resnet} &   & \color[HTML]{9B9B9B}68.6  & \color[HTML]{9B9B9B}87.9 & \color[HTML]{9B9B9B}1.64 & \color[HTML]{9B9B9B}  3.8M  \\
        TSM-R50  \cite{tsm}  & ResNet50  \cite{resnet}  & IN-1K  & \textbf{70.0}    & \textbf{89.5} & \textbf{2.64}     &   24.3M  \\
        \color[HTML]{9B9B9B}MoViNet-A2 \cite{movinets} &&& \color[HTML]{9B9B9B}75.0 & \color[HTML]{9B9B9B}92.3 & \color[HTML]{9B9B9B}1.70 & \color[HTML]{9B9B9B} 4.8M \\ \bottomrule
    \end{tabular}
    }
    \label{tab:k400}
\end{subtable}
\hspace*{\fill}%
\begin{subtable}[t]{.22\linewidth}
    \centering
    \caption{Something Something V2}
    \setlength{\tabcolsep}{3pt} %
    \resizebox{!}{58pt}{%
    \begin{tabular}{@{}llllr@{}}
        \toprule
        Model                & Pretrain   & Top 1         & Top 5         & Latency     \\ \midrule
        NL-R50               & IN-1K      & 47.5          &  78.6         & 4.65        \\
        ViViT-FE             & IN-1K      & 49.9          &  77.0         & 2.95        \\
        Swin-T               & IN-1K      & 52.3          &  80.2         & 3.64        \\
        TEA-R50              & IN-1K      & 52.6          &  80.4         & 4.25        \\
        VTN-EFF              & IN-1K      & 52.9          &  80.6         & \textbf{1.75}   \\
        VTN-R50              & IN-1K      & 53.6          &  80.8         & 2.42        \\
        I3D-BERT             & IN-1K      & 57.8          &  85.2         & 2.49        \\
        TimeSformer          & IN-1K      &\textbf{62.7}  &  89.8         & 10.20       \\ \midrule
        \multicolumn{3}{@{}l}{\textit{non-transformer baselines}} \\
        MoViNet-A0           & K400       & 57.2          & 84.9    & \textbf{1.00} \\
        X3D-XS               & K400       & 58.1          & 85.4          & 1.64        \\
        TSM-R50              & IN-1K      & 59.5          & 86.1          & 2.64        \\ 
        MoViNet-A2           & K400       & \textbf{60.6} &  \textbf{86.5}    & 1.70        \\ \bottomrule
    \end{tabular}
    }
    \label{tab:ssv2}
\end{subtable}
\hspace*{\fill}%
\begin{subtable}[t]{.29\linewidth}
    \centering
    \caption{Epic Kitchens 100 Top 1 Accuracy}
    \setlength{\tabcolsep}{3pt} %
    \resizebox{!}{58pt}{%
    \begin{tabular}{@{}lllllr@{}}
        \toprule
        Model                & Pretrain   & Action        & Verb          & Noun    & Latency   \\ \midrule
        Swin-T               & IN-1K      & 25.9          & 51.0          & 40.5    & 3.64      \\
        NL-R50               & IN-1K      & 28.5          & 52.0          & 45.3    & 4.65      \\
        ViViT-FE             & IN-1K      & 30.3          & 58.4          & 41.3    & 2.95      \\
        TimeSformer          & IN-1K      & 31.3          & 54.6          & \textbf{48.4} & 10.20 \\
        TEA-R50              & IN-1K      & 31.3          & 61.0          & 42.2    & 4.25      \\
        I3D-BERT             & IN-1K      & 31.5          & 62.0          & 41.8    & 2.49      \\
        VTN-EFF              & IN-1K      & 31.9          & 60.9          & 42.2    & \textbf{1.75} \\
        VTN-R50              & IN-1K      & \textbf{33.6} & \textbf{62.5} & 44.3    & 2.42      \\ \midrule
        \multicolumn{3}{@{}l}{\textit{non-transformer baselines}} &      \\
        MoViNet-A0           & K400       & 31.3          & 62.4          & 41.1    & \textbf{1.00} \\
        X3D-XS               & K400       & 32.6          & 62.5          & 43.4    & 1.64      \\
        TSM-R50              & IN-1K      & 32.7          & 60.1          & 44.7    & 2.64      \\ 
        MoViNet-A2           & K400       & \textbf{34.4} & \textbf{64.2} & \textbf{44.8} & 1.70 \\  \bottomrule
    \end{tabular}
    }
    \label{tab:ek100}
\end{subtable}
\vspace{-15pt}
\end{table*}
\vspace{-4pt}
\section{Experiments: Action Recognition}
\label{sec:accuracy}

In the previous section, we have evaluated transformer-based models in terms of efficiency. Next, we benchmark each of them on the task of action recognition, in order to compare their essential latency-accuracy trade-off.

\subsection{Experimental Setup}

\paragraph{Datasets}
We use the Kinetics 400 (K400), Something Something V2 (SSV2), and Epic Kitchens 100 (EK100) datasets \cite{k400, ssv2, ek100}. \Cref{tab:datasets} gives an overview of each dataset. We report top-1 and top-5 classification accuracy (\%) on the validation sets. For EK100, we report top-1 accuracy and additionally report the verb and noun accuracy.

\vspace{-10pt}
\paragraph{Training}
While models are often trained and evaluated under different conditions using different training tricks, we aim for a broadly fair comparison, and thus train all models under the same conditions. It is common to train video models on 64 or 128 GPUs with a batch size of up to 1024 \cite{1hour, slowfast, mvit, x3d, vidtr, motionformer}. However, this is not feasible for the average researcher. We thus restrict our study to a simple and lightweight training strategy. We use the same training strategy across K400, SSV2, and EK100. 

Each model is initialized from publicly available weights pretrained on ImageNet-1K. Using mixed-precision, models are then trained for up to 35 epochs with an effective batch size of 32. Each model uses an initial learning rate between $10^{-4}$ and $3\times10^{-4}$, and a weight decay of $10^{-5}$. We reduce the learning rate by a factor of $0.1\times$ when validation accuracy saturates for 2 epochs. For frame sampling we use the sparse temporal sampling strategy from \cite{tsnraw} to sample 16 RGB frames from a video, which are resized to $224\times224$. The three datasets use slightly different augmentations. Broadly speaking, they are random flipping, random affine (translation, scaling, shearing, rotation), color jittering, and random erasing \cite{erase}. The exact parameters can be found in the supplemental material. Because public ImageNet weights for the X3D and MoViNet baselines are not available, we initialize them from K400 pretrained weights when training on SSV2 and EK100, and simply report their official state-of-the-art results on Kinetics 400.

This simple training environment uses considerably less sophisticated methods and resources than state-of-the-art strategies do. We thus expect that the models we train fall short of top public accuracies. However, our aim is not to squeeze state-of-the-art accuracy out of models, but instead compare all transformer-based models on a level playing field\footnote{{\scriptsize Neither ImageNet nor K400 pretrained weights are publicly available for MViT-S. To avoid an unfair comparison, we do not attempt to pretrain MViT-S in our simplified training environment and instead leave out MViT-S from this section.}}.

\paragraph{Inference}
For each validation video, we evenly extract 16 frames and resize frames to $224\times224$. All models are evaluated under single-crop, single-view inference, \textit{i.e}, models make a single prediction per video.

\subsection{Results}

\paragraph{Kinetics 400}
\Cref{tab:k400} shows that on Kinetics 400, almost all transformer-based models are unable to perform on par with the non-transformer baselines. Compared to the TSM-R50 baseline, Swin-T, ViViT-FE, NL-R50, and TEA-R50 have a 1.00, 0.31, 2.01, and 1.61 higher absolute latency, with a -0.2\%, -1.2\%, -3.1\%, and -4.0\% absolute accuracy deficit. VTN-R50, and I3D-BERT are slightly faster than TSM-R50 by 0.22 and 0.15 absolute latency, yet show a significant accuracy drop of -2.8\% and -5.2\%. The only transformer-based model able to reach an accuracy higher than TSM-R50 is TimeSformer at 74.5\%. While this is a considerable accuracy increase, TimeSformer is by far the slowest of all models with a latency of 10.20. The other attention-only models Swin-T and ViViT-FE attempt to improve the efficiency of the TimeSformer architecture, but are unable to keep the same accuracy as TSM-R50 when reducing the latency. The only transformer-based model able to compete with the baseline TSM-R50 is VTN-EFF, which however moves from a ResNet50 backbone to EfficientNet-B0. Replacing TSM-R50's ResNet50 backbone with the same EfficientNet-B0 backbone likely nullifies the improvement of VTN-EFF over TSM-R50. In summary, based on the Kinetics 400 dataset, \textit{transformer-based models are not as capable of lightweight action recognition as traditional non-transformer models are}. 

\vspace{-8pt}

\paragraph{Something Something V2}
\Cref{tab:ssv2} shows training results on the more motion-heavy SSV2 dataset. In the same way as before, all transformer-based models apart from TimeSformer have similar or higher latency than the TSM-R50 baseline and are unable to match its accuracy. More so, the most efficient MoViNet-A0 baseline has a 3.6\%, 4.3\%, 4.6\%, 4.9\%, 7.3\%, and 9.7\% higher accuracy and is 2.42\texttimes, 1.75\texttimes, 4.25\texttimes, 3.64\texttimes, 2.95\texttimes, and 4.65\texttimes\ faster than VTN-R50, VTN-EFF, TEA-R50, Swin-T, ViViT-FE, and NL-R50, respectively. The only transformer-based models that reach a higher accuracy by 0.6\% and 5.2\% are I3D-BERT and TimeSformer, but do so with a whopping 2.49\texttimes\ and 10.20\texttimes\ increase in latency, respectively. All four non-transformer baselines provide a much more competitive set of latency-accuracy ratios compared to the transformer-based models.

\vspace{-8pt}

\paragraph{Epic Kitchens 100}
\label{par:ek100}
\Cref{tab:ek100} displays the verb, noun, and action classification accuracy of models on the temporally demanding EK100 dataset. Looking at the transformer-based models, there is a clear separation in terms of verb accuracy between the convolution-based models VTN-R50, VTN-EFF, I3D-BERT, and TEA-R50 in the top 4 (with the exception of NL-R50), and the attention-only models ViViT-FE, TimeSformer, and Swin-T. Those built upon convolutions reach an average verb accuracy of 61.6\%, while the attention-only models fall behind by -3.2\%, -7.0\%, and -10.6\%. This suggests that \textit{attention-only models that abolish convolutions are not as capable at modelling motion in videos as convolution-based transformer models}. Furthermore, yet again the transformer-based models are inferior to the non-transformer baselines. X3D-XS has a better latency (1.64) \textit{and} accuracy (32.6\%) than all transformer-based models (1.75 to 10.20 latency and 25.9\% to 31.9\% accuracy), with the exception of VTN-R50, which has a 1.0\% higher accuracy, but still an approximately 50\% higher latency.

\subsection{Putting It All Together}

\begin{figure}[]
  \centering
  \scriptsize
  \includegraphics[width=0.81\linewidth]{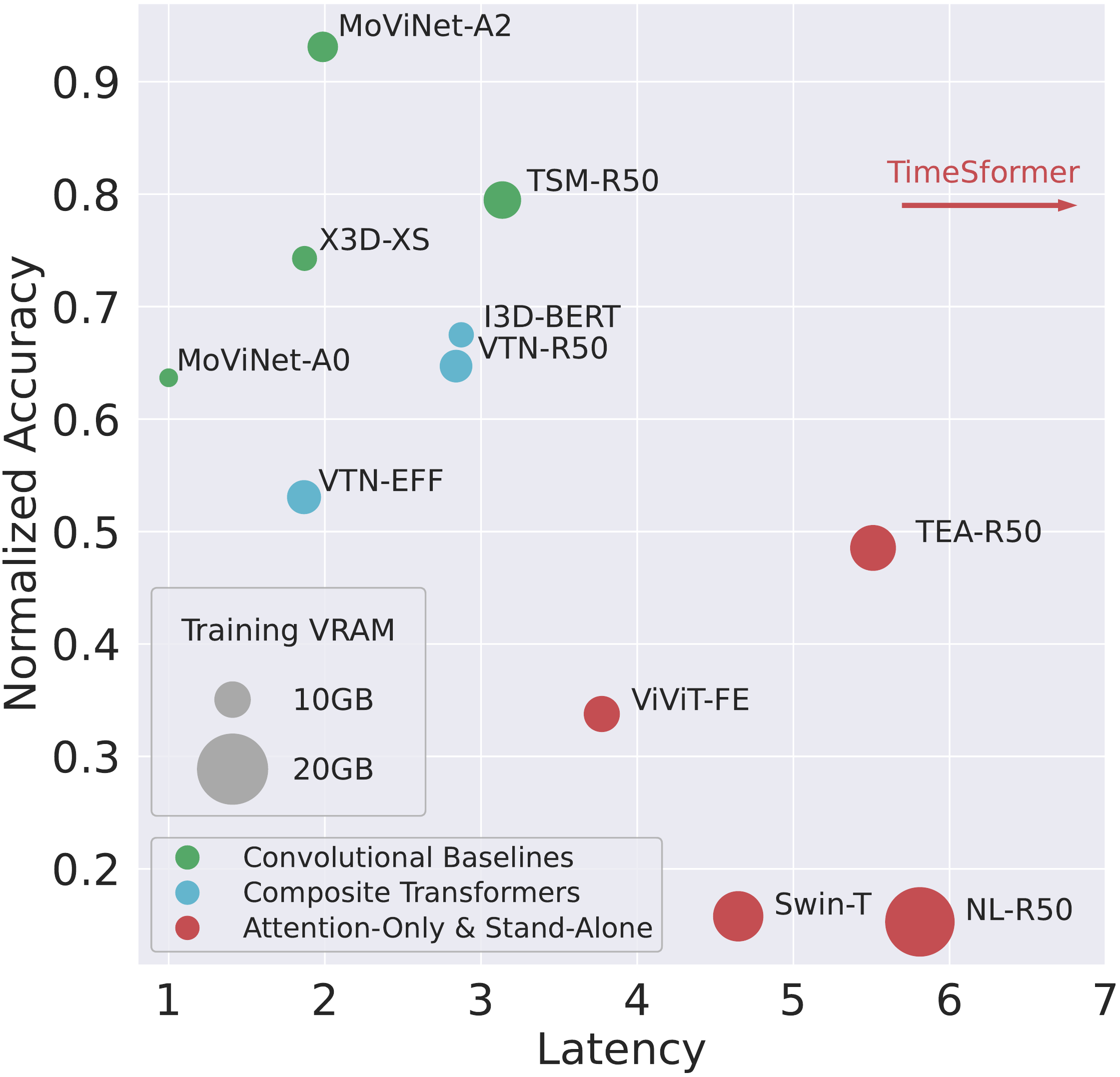}
  \vspace{-10pt}
  \caption[Caption for Figure 5]{A summary of all previous experiments (best viewed in color). The non-transformer baselines in green clearly outperform the transformer-based methods. Latency is the average over training, batch-mode inference, and single-instance inference latency. Accuracy is the average of min-max normalized accuracy over the SSV2 and EK100 datasets.}
  \label{fig:summary}
\vspace{-15pt}
\end{figure}

We summarize all experiments in \cref{fig:summary}. For latency, we average training, batch-mode inference, and single-instance inference latency from \Cref{fig:train_latency,,fig:bm_inference_latency,,fig:si_inference_latency}. For accuracy, we min-max normalize the accuracies of each dataset, and then average them across models, resulting in a single summary statistic for performance (excluding K400 due to differing training strategies).

Overall, traditional convolutional baselines (green) reach a much better latency-accuracy ratio than any other transformer-based methods. Within the subset of transformer-based methods, composite transformers (blue) based on convolutions are much more capable of lightweight action recognition than models based on attention-only architectures and stand-alone attention blocks (red). Based on the EK100 benchmark in \cref{par:ek100}, we have seen that attention-only models (red) are significantly worse at motion modelling than composite transformers (blue). In the latency benchmarks from \Cref{sec:training_latency}, we have also seen that stand-alone attention blocks (red) cause a significant latency overhead. Ultimately, Composite transformers (blue) seem to avoid both deficits, by only adding small transformers to convolutional backbones with better inductive biases. 

\vspace{-5pt}
\section{Limitations and Future Directions}
\vspace{-3pt}
In-spite of our efforts to conduct a fully fair comparison, there are a number of limitations. While we evaluate all models under the same training and testing conditions to avoid unfair advantages, our chosen training and testing strategy might be more optimal for some models than for others. Certain additional data augmentations \cite{cutmix, mixup, randaug, videomix}, training schedules \cite{1hour}, or regularization strategies \cite{deit, labelsmooth}, might make individual models underperform less. Furthermore, for 3 out of 4 baselines on SSV2 and EK100, we use Kinetics 400 pretraining instead of ImageNet-1K. While both Kinetics 400 and ImageNet-1K are similarly large-scale and static pretraining datasets, cross-comparing models pretrained on either is potentially less fair. However, on every dataset, an ImageNet-1K pretrained TSM model counters this. Lastly, our study is based on efficiency evaluations in the form of real-world latency measurements, instead of theoretical FLOPs. Choosing either has its disadvantages. While FLOPs is hardware-agnostic, it is not an accurate representation of real world runtimes. At the same time, while latency gives a much more accurate measure of efficiency, it runs the risk of changing over time due to hardware developments and CUDA optimizations.

In spite of these limitations, our study compares models as robustly as possible and avoids many fairness pitfalls often present in other works. Our study is the first to evaluate the efficiency of lightweight action recognition models in depth across multiple devices and computation modes, and the first to train a wide range of video transformers under the same conditions.

Our study helps better inform future research. A promising direction is improving the motion modeling capability of attention-only models. Regarding efficiency, MViT has shown that the inner dimensionality of self-attention is a much more promising target for improving transformer efficiency, than progressive downsampling, encoder factorization, attention factorization, or linear self-attention. Stand-alone attention blocks also show a considerable latency overhead, which could be worth improving. Strict attention-only has shown to be insufficient for lightweight action recognition in general. So, finding new ways to combine these architectures with other paradigms, like composite transformers do, is a promising direction. The inconsistency and complexity of our mobile latency results also highlight the need for in-depth studies on mobile video transformers. At last, other works have shown that transformer models in vision often require more data or more sophisticated training strategies \cite{vit, deit}. Therefore, it is worth exploring whether repeating our latency and accuracy study with larger-scale pretraining and more sophisticated training strategies changes the landscape of our results.

\vspace{-5pt}
\section{Conclusion}
\vspace{-3pt}
In this work, we have asked the question which video transformers are most efficient and whether they are as capable of lightweight action recognition as traditional baselines are. After comparing 10 video transformers to 3 traditional convolutional baselines, we have concluded that more work is needed for video transformers to match the latency-accuracy ratio of traditional baselines. Key research directions include better composing transformers with other lightweight paradigms, improving the motion modeling ability of attention-only architectures, and reducing the latency overhead of stand-alone attention blocks. Video transformers on mobile devices also warrant additional exploration to find the consistently best approach. Future research in these directions has the potential of making video transformers much more viable for lightweight action recognition.

\appendix
\captionsetup[figure]{font=normalsize}
\captionsetup[table]{font=normalsize}
\section*{Appendix}

\section{FLOPs}
\label{sec:flops}
\begin{figure}[h]
  \centering
  \includegraphics[width=1.0\linewidth]{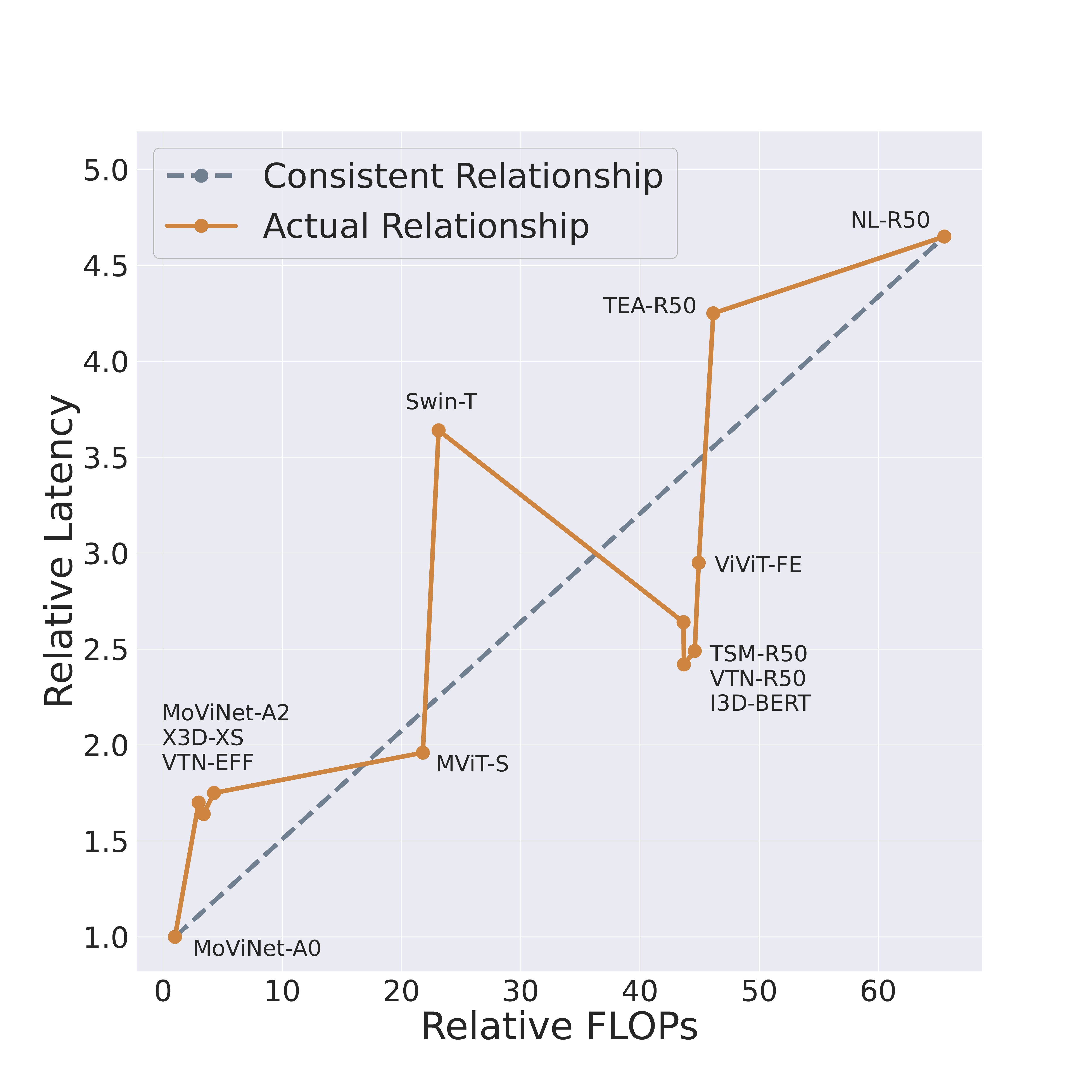}

  \caption[Caption for Figure 1]{A comparison of FLOPs and training latency (excluding TimeSformer).}
  \label{fig:flops}
\end{figure}

\Cref{fig:flops} shows the relationship between the FLOPs and latency of a model. The dashed grey line shows the expected consistent relationship. The brown line shows the actual empirical relationship, based on our latency measurements. The inconsistency of the brown line shows that FLOPs is not a reliable measure of real-world runtime. Models that have almost identical FLOPs, such as MViT-S and Swin-T, have a wildly different latency. Also, according to FLOPs, the slowest model NL-R50 is over 60\texttimes\ slower than the fastest model MoViNet-A0. However, in reality
NL-R50 is only 4.65\texttimes\ slower than MoViNet-A0. This shows that to accurately compare the real-world runtime of models, FLOPs can not be relied upon. Instead, latency must be measured, as in our experiments.

\begin{table}[]
\centering
\caption{Comparison of the number of training epochs required for each model to converge on SSV2 \cite{ssv2}. Latency is average training latency. VRAM is the video memory required during training using a batch size of 8, 16 frames, and a frame size of $224 \times 224$. Latency \texttimes\ Epochs shows that some models take much less or more overall training time than other models.}
\label{tab:training}
\resizebox{1.0\linewidth}{!}{%
\begin{tabular}{@{}lrrrc@{}}
\toprule
Model & Latency & VRAM & Epochs & Latency \texttimes\ Epochs \\ \midrule
TimeSformer \cite{timesformer}      &   10.20       &   22.2G       & 9  &  91.8    \\
NL-R50   \cite{non-local}           &   4.65        &   19.5G       & 21 &  97.7 \\
TEA-R50   \cite{tea}                &   4.25        &   12.7G       & 15 &  63.8 \\
Swin-T   \cite{videoswin}           &   3.64        &   14.0G       & 9  &  32.8              \\
ViViT-FE  \cite{vivit}              &   2.95        &   9.9G        & 11 &  32.5           \\
I3D-BERT \cite{videobert}           &   2.49        &   6.8G        & 15 &  37.4        \\
VTN-R50  \cite{vtn}                 &   2.42        &   8.9G        & 15 &  36.3         \\
VTN-EFF \cite{vtn}                  &\textbf{1.75}  &   9.3G        & 20 &  35.0        \\ 
\midrule
\multicolumn{5}{@{}l}{\textit{non-transformer baselines}}               \\
TSM-R50  \cite{tsm}                 &   2.64        &   10.3G       & 17 & 44.9         \\
MoViNet-A2 \cite{movinets}          &   1.70        &   8.3G        & 10 & 17.0          \\
X3D-XS  \cite{x3d}                  &   1.64        &   6.7G        & 9  & 14.8     \\
MoViNet-A0 \cite{movinets}          &\textbf{1.00}  &\textbf{4.9G}  & 10 & 10.0   \\
\bottomrule
\end{tabular}
}
\end{table}

\begin{table}[]
\centering
\caption{The starting learning rate of each model and whether the ImageNet/Kinetics \cite{imagenet, k400} pretrained backbone was frozen for the first epoch.}
\label{tab:learning_rate}
\resizebox{0.75\linewidth}{!}{%
\begin{tabular}{@{}lcc@{}}
\toprule
Model & Learning Rate & Freeze \\ \midrule
TimeSformer \cite{timesformer}      &   $1\times10^{-4}$        &           \\
NL-R50   \cite{non-local}           &   $1\times10^{-4}$        &  \cmark         \\
TEA-R50   \cite{tea}                &   $1\times10^{-4}$        &         \\
Swin-T   \cite{videoswin}           &   $1\times10^{-4}$        &  \cmark                  \\
ViViT-FE  \cite{vivit}              &   $1\times10^{-4}$        &  \cmark               \\
I3D-BERT \cite{videobert}           &   $2\times10^{-4}$        &  \cmark              \\
VTN-R50  \cite{vtn}                 &   $2\times10^{-4}$        &  \cmark               \\
VTN-EFF \cite{vtn}                  &   $2\times10^{-4}$        &  \cmark            \\ 
\midrule
\multicolumn{3}{@{}l}{\textit{non-transformer baselines}}               \\
TSM-R50  \cite{tsm}                 &   $1\times10^{-4}$        &  \cmark              \\
MoViNet-A2 \cite{movinets}          &   $2\times10^{-4}$        &  \cmark               \\
X3D-XS  \cite{x3d}                  &   $2\times10^{-4}$        &  \cmark           \\
MoViNet-A0 \cite{movinets}          &   $2\times10^{-4}$        &  \cmark    \\
\bottomrule
\end{tabular}
}
\end{table}

\clearpage
\section{Training Hyperparameters}
\label{sec:parameters}
\Cref{tab:learning_rate} shows the initial learning rate setting of each model during training. \Cref{sec:ek100,,sec:ssv2,,sec:k400} show the data augmentations used for each dataset. Parameter values correspond to Torchvision \cite{pytorch} transform parameters.

\subsection{Epic Kitchens 100}
\label{sec:ek100}
\begin{enumerate}
  \item Resize to $224\times224$
  \vspace{-8pt}
  \item Random Horizontal Flip
  \vspace{-8pt}
  \item Random Vertical Flip
  \vspace{-8pt}
  \item Random Affine \\{\small(degrees=15, translate=0.15, scale=[0.85, 1.15], shear=+-15)}
  \vspace{-15pt}
  \item ColorJitter \\{\small(brightness=0.35, contrast=0.35, saturation=0.2, hue=0.1)}
  \vspace{-3pt}
  \item Random Erasing \\{\small(p=0.7, scale=[0.05, 0.1])}
  \vspace{-3pt}
  \item Normalize
\end{enumerate}

\subsection{Something Something V2}
\label{sec:ssv2}
\begin{enumerate}
  \item Resize to $224\times224$
  \vspace{-8pt}
  \item Random Affine \\{\small(degrees=15, translate=0.15, scale=[0.85, 1.15], shear=+-15)}
  \vspace{-15pt}
  \item ColorJitter \\{\small(brightness=0.35, contrast=0.35, saturation=0.2, hue=0.1)}
  \vspace{-3pt}
  \item Random Erasing \\{\small(p=0.7, scale=[0.05, 0.1])}
  \vspace{-3pt}
  \item Normalize
\end{enumerate}

\subsection{Kinetics 400}
\label{sec:k400}
\begin{enumerate}
  \item Resize to $224\times224$
  \vspace{-8pt}
  \item Random Horizontal Flip
  \vspace{-8pt}
  \item Random Affine \\{\small(degrees=10, translate=0.15, scale=[0.85, 1.15], shear=+-15)}
  \vspace{-15pt}
  \item ColorJitter \\{\small(brightness=0.35, contrast=0.35, saturation=0.2, hue=0.1)}
  \vspace{-3pt}
  \item Random Erasing \\{\small(p=0.7, scale=[0.05, 0.1])}
  \vspace{-3pt}
  \item Normalize
\end{enumerate}

\section{Latency Benchmarks}
\begin{figure}[]
  \centering
  \includegraphics[width=1.0\linewidth]{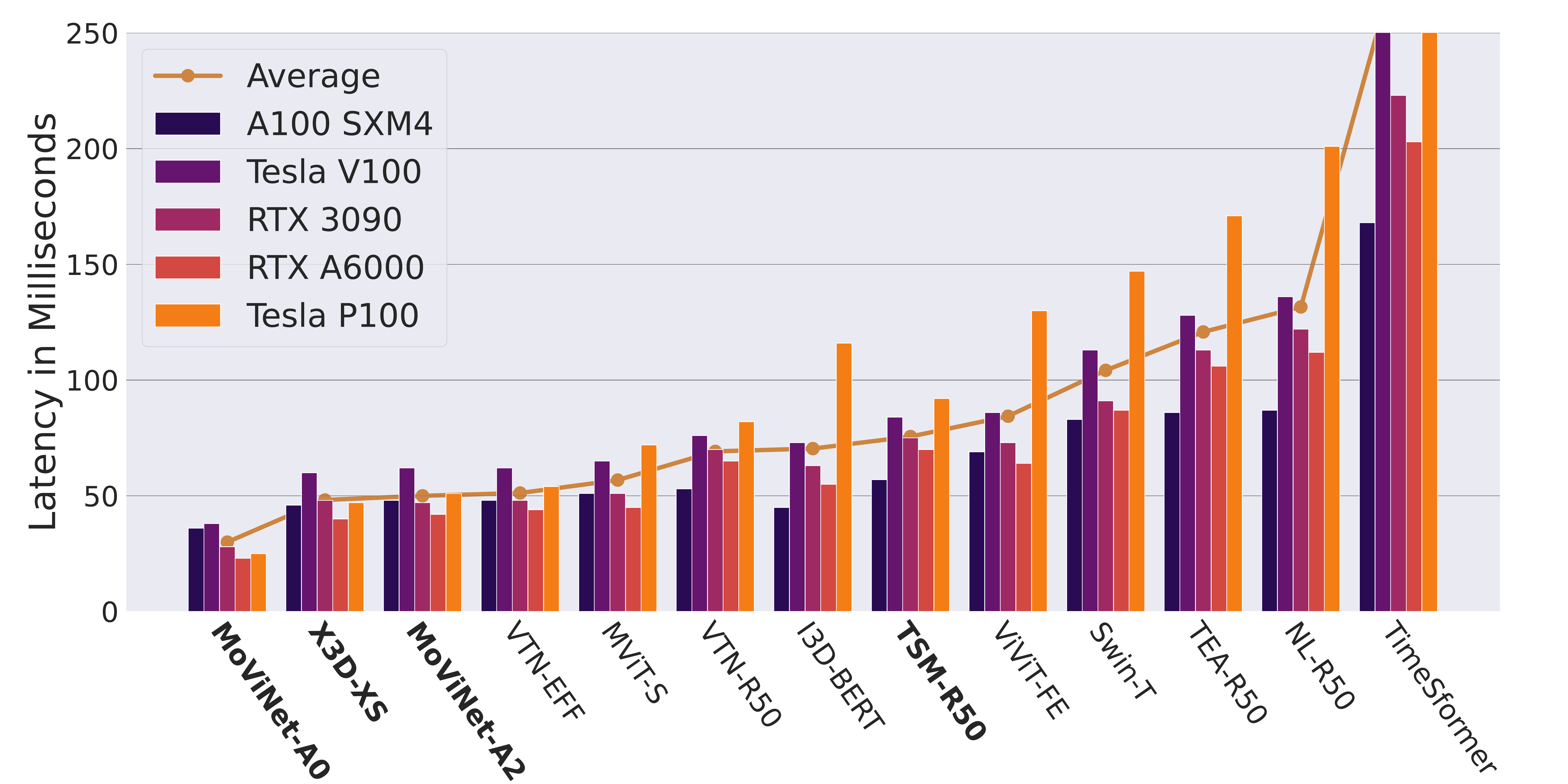}
  \caption[Caption for Figure 1]{Absolute training step latency of models on different desktop GPUs (best viewed in color). Baselines are in bold.}
  \label{fig:train_latency_absolute}
\end{figure}

\begin{figure}[]
  \centering
  \includegraphics[width=1.0\linewidth]{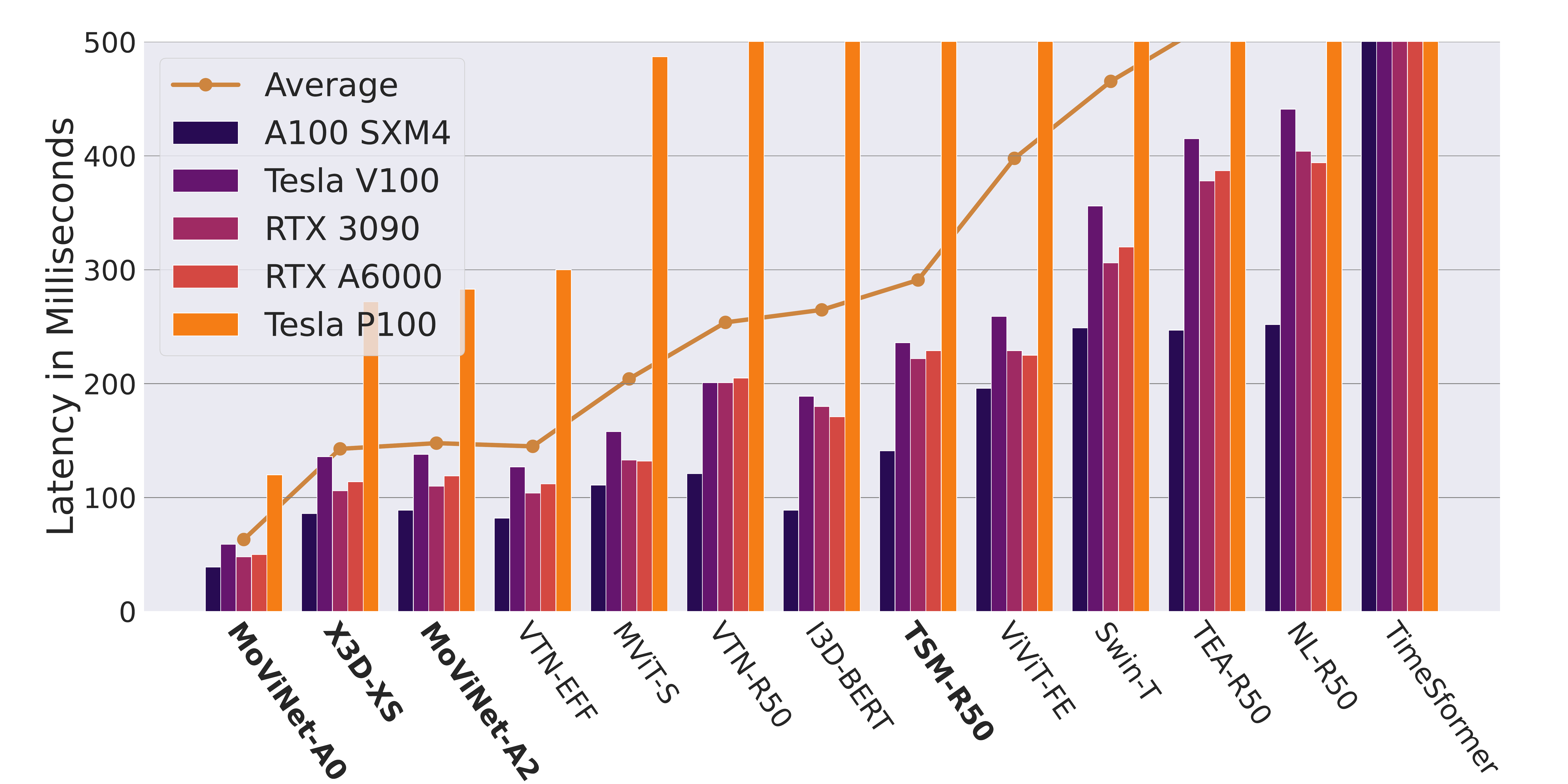}
  \caption[Caption for Figure 1]{Absolute \textit{batch-mode} inference latency of models on different desktop GPUs (Best viewed in color). Baselines are in bold.}
  \label{fig:bm_latency_absolute}
\end{figure}

\begin{figure}[]
  \centering
  \includegraphics[width=1.0\linewidth]{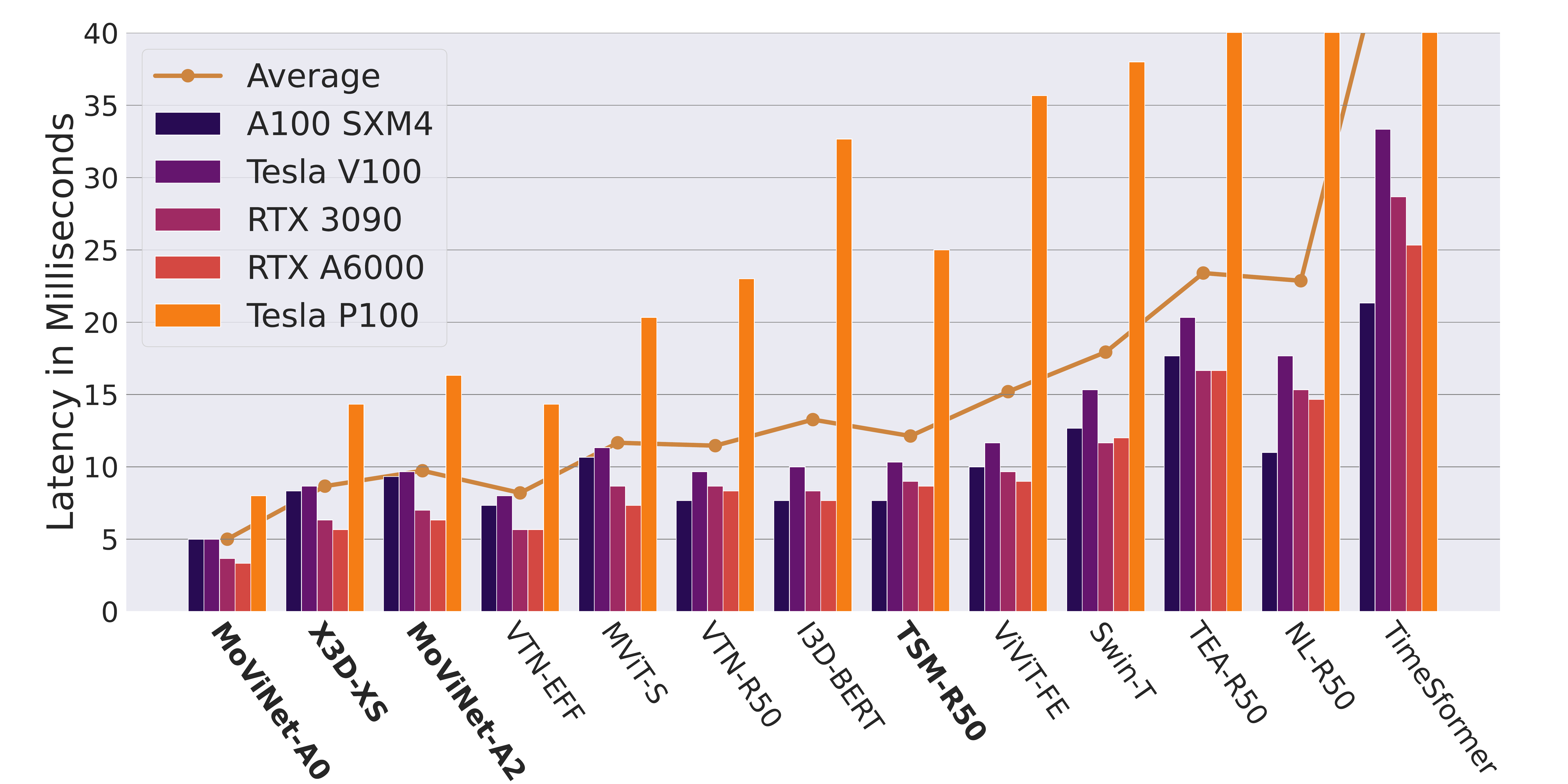}
  \caption[Caption for Figure 1]{Absolute \textit{single-instance} inference latency of models on different desktop GPUs (Best viewed in color). Baselines are in bold.}
  \label{fig:si_latency_absolute}
\end{figure}

All desktop GPU benchmarks are run using PyTorch 1.10, Torchvision 0.11.1, CUDA 11.3, and cuDNN 8.2.0. \Cref{fig:train_latency_absolute,,fig:bm_latency_absolute,,fig:si_latency_absolute} report absolute training and inference latency. 

\begin{figure}
    \centering
    \begin{subfigure}[b]{0.32\linewidth}
        \centering
        \includegraphics[width=1.0\linewidth]{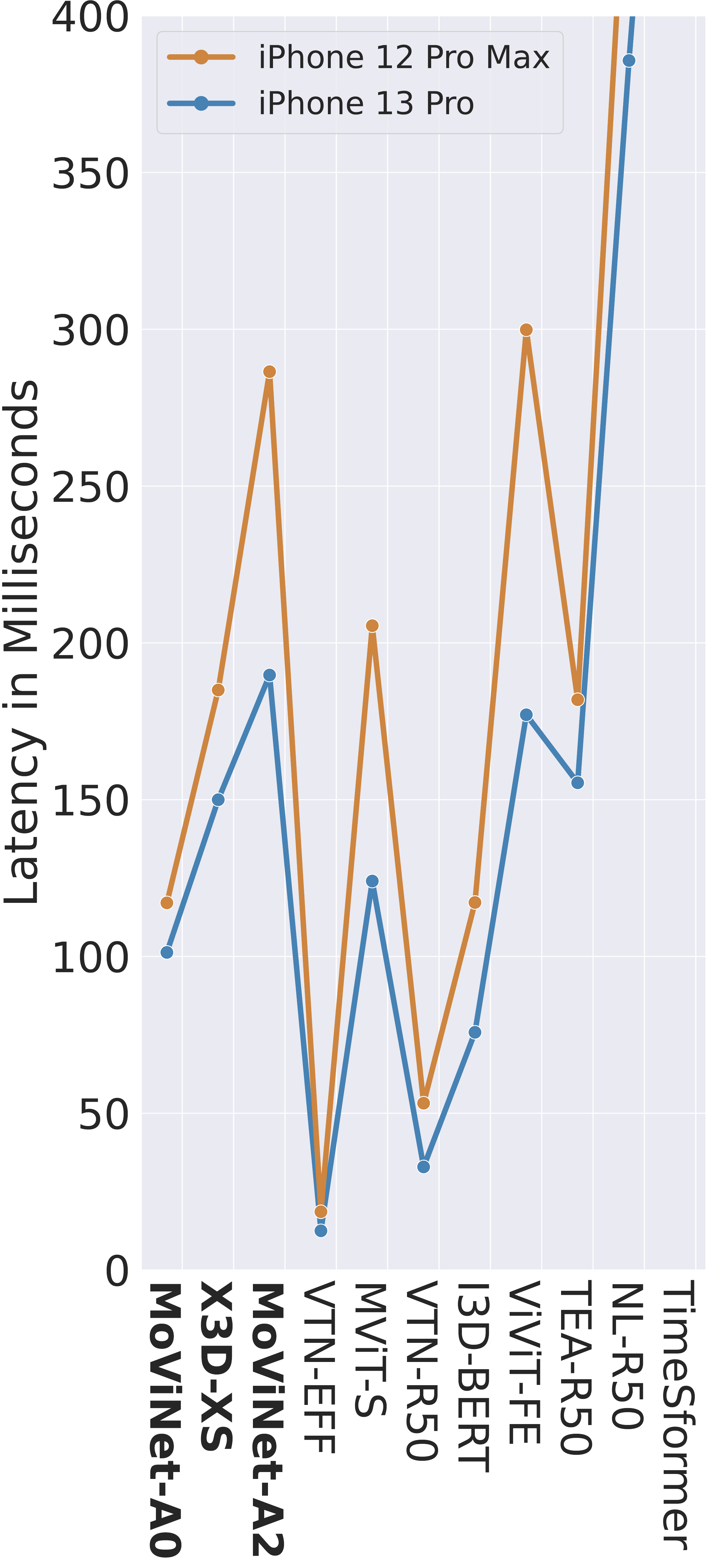}
        \label{fig:ane_mobile_absolute}
        \caption{ANE}
    \end{subfigure}
    \hfill
    \begin{subfigure}[b]{0.32\linewidth}
        \centering
        \includegraphics[width=1.0\linewidth]{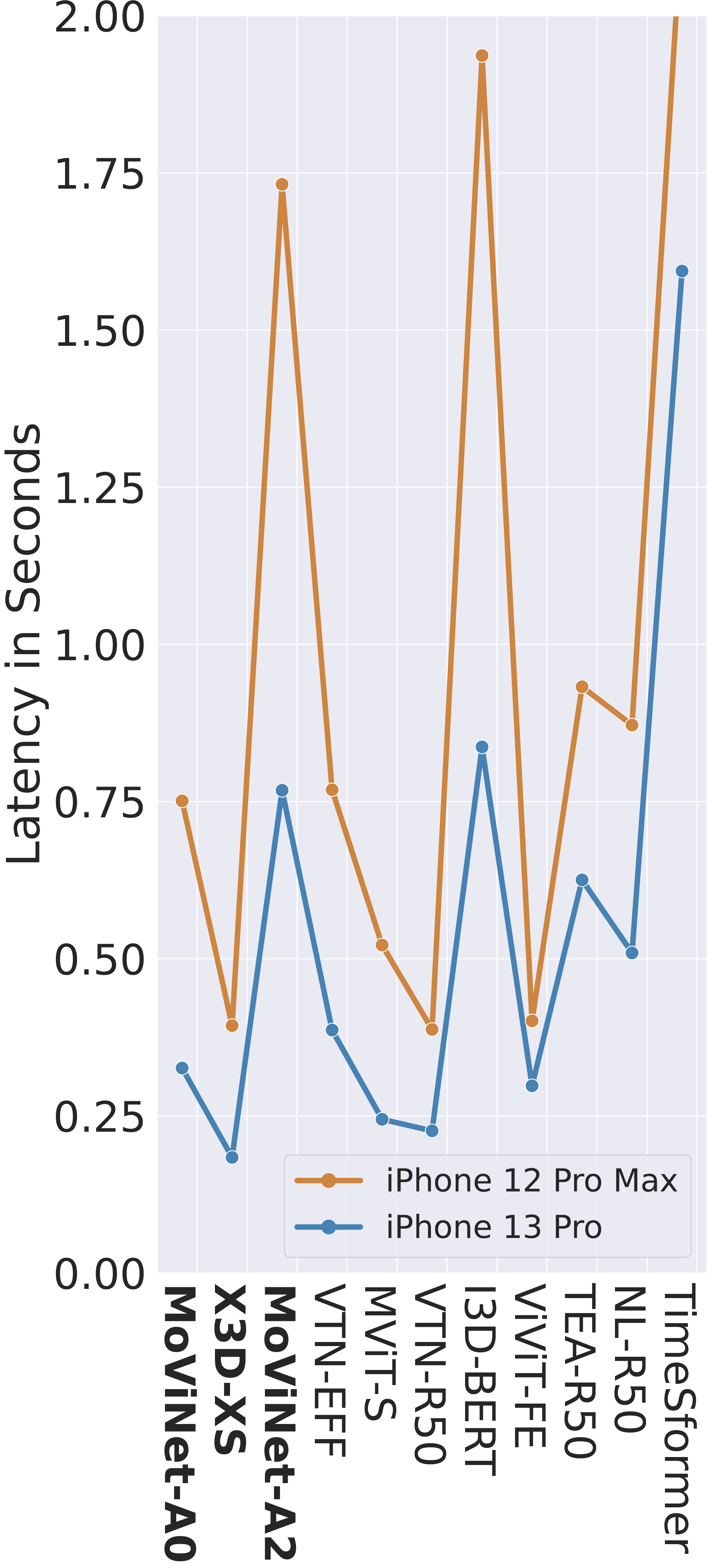}
        \label{fig:gpu_mobile_absolute}
        \caption{GPU}
    \end{subfigure}
    \hfill
    \begin{subfigure}[b]{0.32\linewidth}
        \centering
        \includegraphics[width=1.0\linewidth]{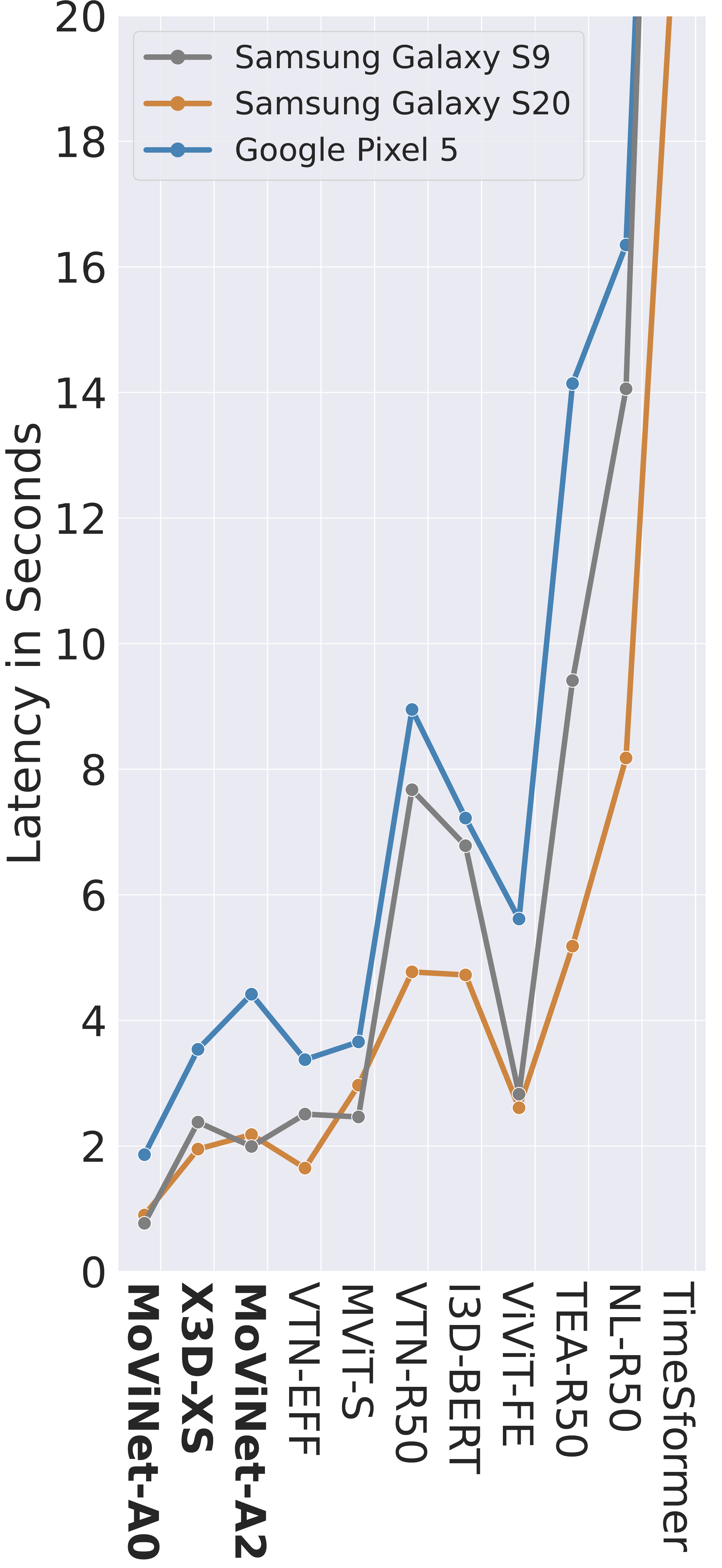}
        \label{fig:cpu_mobile_absolute}
        \caption{CPU}
    \end{subfigure}
    \caption{Absolute \textit{single-instance mobile} inference latency on the CPU, GPU, and ANE accelerator. Note the different y-axis scales.}
    \label{fig:mobile_latency_absolute}
\end{figure}

\cref{fig:mobile_latency_absolute} reports absolute mobile inference latency. ANE latency, which is given in milliseconds, is significantly faster than GPU latency on the same devices. On the CPU of Android devices, most models are extremely slow. Only a few models, such as the baselines and VTN-EFF are able to run in under 4 seconds or even under 2 seconds.

\vspace{30pt}

\vspace{30pt}

{\small
\vspace{10pt}
\bibliographystyle{ieee_fullname}
\bibliography{egbib}
}

\end{document}